\documentclass{article}
\usepackage{spconf,amsmath,graphicx}

\usepackage[colorlinks=true,allcolors=black]{hyperref}
\usepackage{url}
\usepackage{textcomp} 
\usepackage{gensymb} 
\usepackage{arydshln}
\usepackage{booktabs}
\usepackage{multirow}
\usepackage[utf8]{inputenc}
\usepackage[acronym]{glossaries}
\usepackage{subfigure}
\usepackage[table,dvipsnames]{xcolor}
\usepackage{xspace} 
\usepackage[sort&compress,numbers]{natbib}
\usepackage{balance}

\usepackage{diagbox}

\newcommand\major[1]{#1}
\newcommand\minor[1]{#1}

\usepackage{pgf}
\usepackage{multirow}
\usepackage{tikz}

\usepackage{pgfplots}
\usepackage{pgfplotstable}
\pgfplotsset{compat=1.7}

\usetikzlibrary{calc}
\usetikzlibrary{decorations.pathmorphing}
\usetikzlibrary {patterns,shapes,shadows,decorations.pathreplacing}

\usetikzlibrary{arrows,shadows,positioning}
\usetikzlibrary{arrows.meta}
\usepackage[linesnumbered,ruled,norelsize]{algorithm2e}
\usepackage{flushend}
\usepackage[normalem]{ulem}
\usepackage{textcomp}
\def\BibTeX{{\rm B\kern-.05em{\sc i\kern-.025em b}\kern-.08em
    T\kern-.1667em\lower.7ex\hbox{E}\kern-.125emX}}

\usetikzlibrary{matrix,positioning}
\usetikzlibrary{decorations.markings}
\usetikzlibrary{calc,trees,positioning,arrows,fit,shapes,calc}

\usepackage[export]{adjustbox}

\usepackage{amssymb}
\usepackage{pifont}
\newcommand{\cmark}{\ding{51}}%
\newcommand{\xmark}{\ding{55}}%

\tikzset{
  lrectangle/.style n args={6}{%
    rectangle,
    thick,
    draw,
    fit={(#1,#2) (#3,#4)},
    append after command={\pgfextra{\let\mainnode=\tikzlastnode}
      node[above,inner sep=1pt] at (\mainnode.north) {\scriptsize #5}%
      node[below,inner sep=1pt] at (\mainnode.south) {\tiny #6}%
    },
  }
}

\newcommand{\etal}{\mbox{\emph{et al.}}\xspace}

\tikzset{pics/fake box/.style args={
#1 with dimensions #2 and #3 and #4 and #5 and #6 and #7}{
    code={
      \draw[gray,ultra thin,fill=#1] (0,0,0) coordinate(-front-bottom-left) to ++ (0,#3,0) coordinate(-front-top-right) --++ (#2,0,0) coordinate(-front-top-right) --++ (0,-#3,0) coordinate(-front-bottom-right) node[right,color=black,midway,inner sep=0pt] {} -- cycle;
       
      \draw[gray,ultra thin,fill=#1] (0,#3,0) --++  (0,0,#4) coordinate(-back-top-left) --++ (#2,0,0)  coordinate(-back-top-right) --++ (0,0,-#4)  -- cycle;
       
      \draw[gray,ultra thin,fill=#1] (0,#3,0) --++  (0,0,0.75*#4) --++ (#2,0,0) --++ (0,0,0.75*-#4)  -- cycle;
       
      \draw[gray,ultra thin,fill=#1] (0,#3,0) --++  (0,0,0.50*#4) --++ (#2,0,0) --++ (0,0,0.50*-#4)  -- cycle;
       
      \draw[gray,ultra thin,fill=#1] (0,#3,0) --++  (0,0,0.25*#4) --++ (#2,0,0) --++ (0,0,0.25*-#4)  -- cycle;
     
      \draw[gray,ultra thin,fill=#1!75!black] (#2,0,0) --++ (0,0,#4) coordinate(-back-bottom-right) node[right,color=black,midway,inner sep=0pt] {} --++ (0,#3,0) --++ (0,0,-#4)  -- cycle;
       
      \draw[gray,ultra thin,fill=#1!75!black] (#2,0,0) --++ (0,0,0.75*#4) --++ (0,#3,0) --++ (0,0,0.75*-#4)  -- cycle;
       
      \draw[gray,ultra thin,fill=#1!75!black] (#2,0,0) --++ (0,0,0.5*#4) --++ (0,#3,0) --++ (0,0,0.5*-#4)  -- cycle;
       
      \draw[gray,ultra thin,fill=#1!75!black] (#2,0,0) --++ (0,0,0.25*#4) --++ (0,#3,0) --++ (0,0,0.25*-#4)  -- cycle;
       
      \draw[gray,ultra thin,fill=#1!90!black] (0,0,#4) --++ (0,#3,0) --++ (#2,0,0) node[above,color=black,midway,inner sep=1pt] {} --++ (0,-#3,0) -- cycle; 
    }
  }
}

\tikzset{
        line/.style={
            draw, ->, rounded corners = 3mm,
        }
}

\tikzset{
    *|/.style={
        to path={
            (perpendicular cs: horizontal line through={(\tikztostart)},
                                 vertical line through={(\tikztotarget)})
            -- (\tikztotarget) \tikztonodes
        }
    }
}

\newacronymstyle{long-short-br}
{%
  \GlsUseAcrEntryDispStyle{long-short}%
}%
{%
  \GlsUseAcrStyleDefs{long-short}%
}
\setacronymstyle{long-short-br}

\title{\LARGE \dataset: A New Dataset to Explore Feature Fusion for\\Vehicle Identification Using Convolutional Neural Networks}

\makeatletter
\def\@name{ \emph{Icaro O. de Oliveira\textsuperscript{1}, Rayson Laroca\textsuperscript{2}, David Menotti\textsuperscript{2}},\\\emph{Keiko V. O. Fonseca\textsuperscript{1}, Rodrigo Minetto\textsuperscript{1}}\thanks{
This is an author-prepared version. The published version is available at the \textit{IEEE Xplore Digital Library} (DOI: \href{http://doi.org/10.1109/ACCESS.2021.3097964}{\textcolor{blue}{10.1109/ACCESS.2021.3097964}}).} \\}
\makeatother

\address{\\[-15pt]\textsuperscript{1} Federal University of Technology - Paran\'{a}, Curitiba, Brazil\\\textsuperscript{2} Federal University of Paran\'a, Curitiba, Brazil\\[0.5ex]}

\begin{document}
\sloppy
\newacronym{brnn}{BRNN}{Bidirectional Recurrent Neural Network}
\newacronym{cnn}{CNN}{Convolutional Neural Network}
\newacronym{ctc}{CTC}{Connectionist Temporal Classification}
\newacronym{denatran}{DENATRAN}{National Traffic Department of Brazil}
\newacronym{fps}{FPS}{frames per second}
\newacronym{gan}{GAN}{Generative Adversarial Network}
\newacronym{hog}{HOG}{Histogram of Oriented Gradients}
\newacronym{lstm}{LSTM}{Long Short-Term Memory}
\newacronym{ocr}{OCR}{Optical Character Recognition}
\newacronym{stn}{STN}{Spatial Transform Network}

\newcommand{\citychallenge}{CityFlowV2\xspace}
\newcommand{\dataset}{Vehicle-Rear\xspace}
\newcommand{\verisete}{VeRi-$776$\xspace}
\newcommand{\veriwild}{VERI-Wild\xspace}
\newcommand{\vehicleid}{VehicleID\xspace}

\maketitle
\begin{abstract}
\textit{This work addresses the problem of vehicle identification through non-overlapping cameras. 
\minor{As our main contribution, we introduce a novel dataset for vehicle identification, called Vehicle-Rear, that contains more than three hours of high-resolution videos,
with accurate information about the make, model, color and~year of nearly 3{,}000 vehicles, in addition to the position and identification of their license plates.}
To explore our dataset we design a two-stream \gls*{cnn} that simultaneously uses two of the most distinctive and persistent features available: the vehicle's appearance and its license plate. This is an attempt to tackle a major problem: false alarms caused by vehicles with similar designs or by very close license plate identifiers.
In the first network stream, shape similarities are identified by a Siamese \gls*{cnn} that uses a pair of low-resolution vehicle patches recorded by two different cameras. In the second stream, we use a \gls*{cnn} for \gls*{ocr} to extract textual information, confidence scores, and string similarities from a pair of high-resolution license plate patches.
Then, features from both streams are merged by a sequence of fully connected layers for decision.
In our experiments, we compared the two-stream network against several well-known CNN architectures using single or multiple vehicle features.
The architectures, trained models, and dataset are publicly available at
{\textit{\url{https://github.com/icarofua/vehicle-rear}}}}.
\end{abstract}
%
%
\section{Introduction}~\label{cha:intro}

\glsresetall

Identifying vehicles through non-overlapping cameras is an important task to assist surveillance activities such as travel time estimation, enforcement of speed limits, criminal investigations, and traffic flow. The vehicle identification problem can be formally defined as the process of assigning the same label to distinct instances of the same object as it moves over time in a network of non-overlapping cameras~\cite{bedagkargala2014survey}. 
The remarkable progress of emerging technologies in producing low-cost cameras, capable of acquiring high-definition images, has made the infrastructure to tackle this problem become pervasive in many cities.

\begin{figure}[!htb]
\centering

\begin{tikzpicture}[scale=0.9, every node/.style={scale=0.9}]

\draw(0.0,6.0) node[text centered, text width=3.8cm] (car1) {
{\textbf{Camera 1}}\\
\vspace{0.1cm}
\includegraphics[width=3.8cm,height=3.2cm]{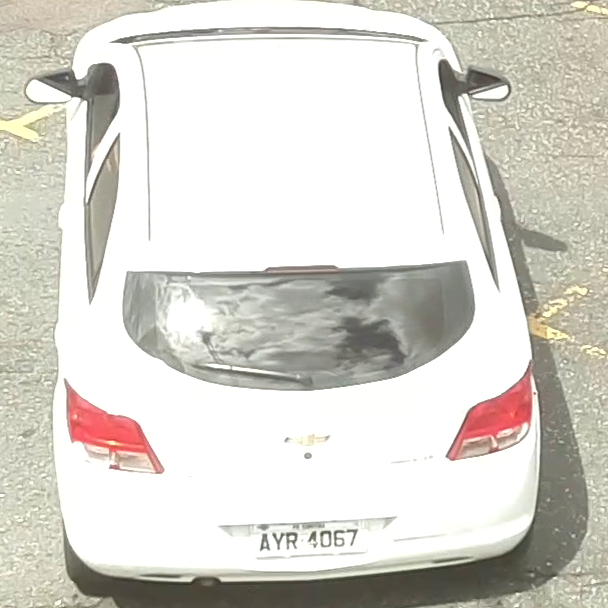}
};

\draw(2.1,4.0) node[text centered, text width=6.8cm] {
\footnotesize (a)
};

\draw(0.8,6.85) node[] (car1) {
 \frame{\includegraphics[width=2.0cm,height=1.0cm]{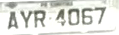}}
};

\draw(4.2,6.0) node[text centered, text width=3.8cm] (car2) {
{\textbf{Camera 2}}\\
\vspace{0.1cm}
\includegraphics[width=3.8cm,height=3.2cm]{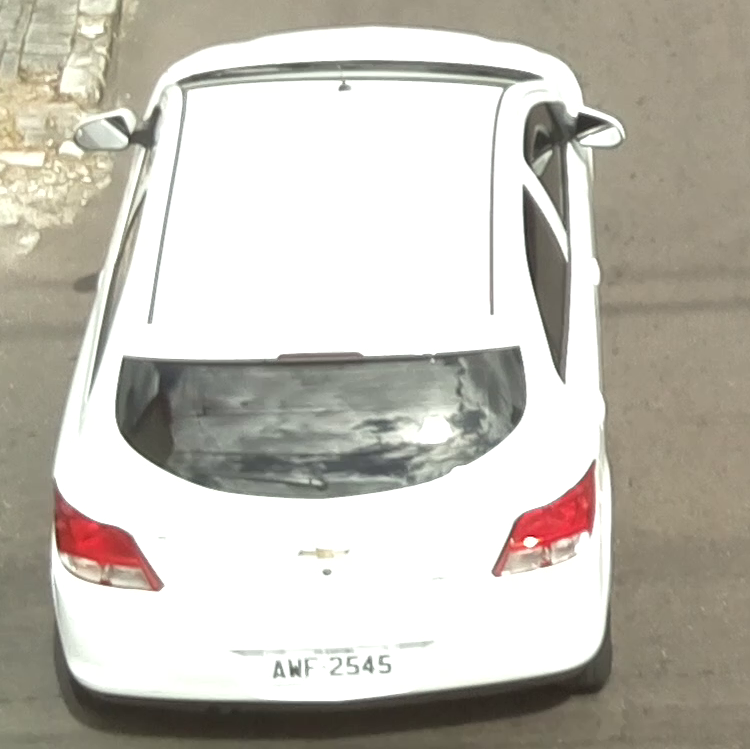}
};

\draw(5.0,6.85) node[] (car1) {
 \frame{\includegraphics[width=2.0cm,height=1.0cm]{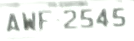}}
};
\end{tikzpicture}
\vspace{0pt}

\begin{tikzpicture}[scale=0.8, every node/.style={scale=0.8}]

\draw(0.0,6.0) node[] (car1) {
\includegraphics[width=3.8cm,height=3.2cm]{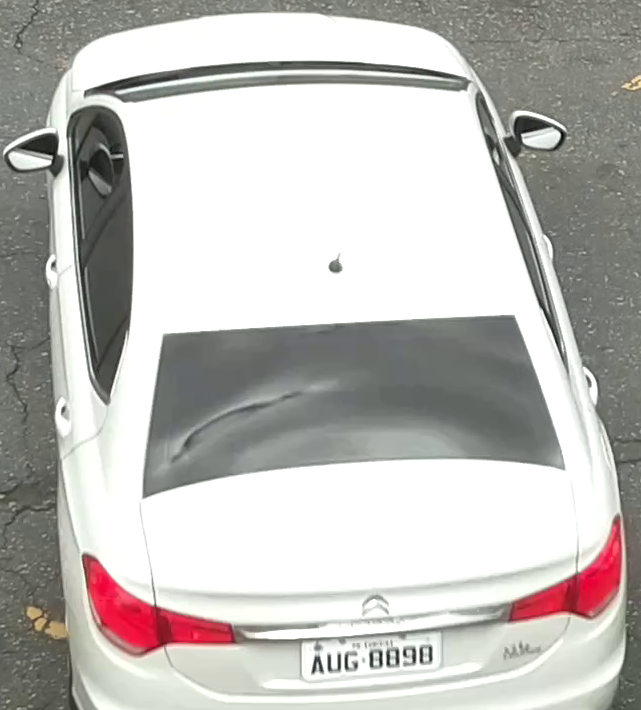}
};

\draw(2.1,4.13) node[text centered, text width=6.8cm] {
\footnotesize (b)
};

\draw(0.8,7.00) node[] (car1) {
 \frame{\includegraphics[width=2.0cm,height=1.0cm]{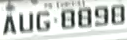}}
};

\draw(4.2,6.0) node[] (car2) {
\includegraphics[width=3.8cm,height=3.2cm]{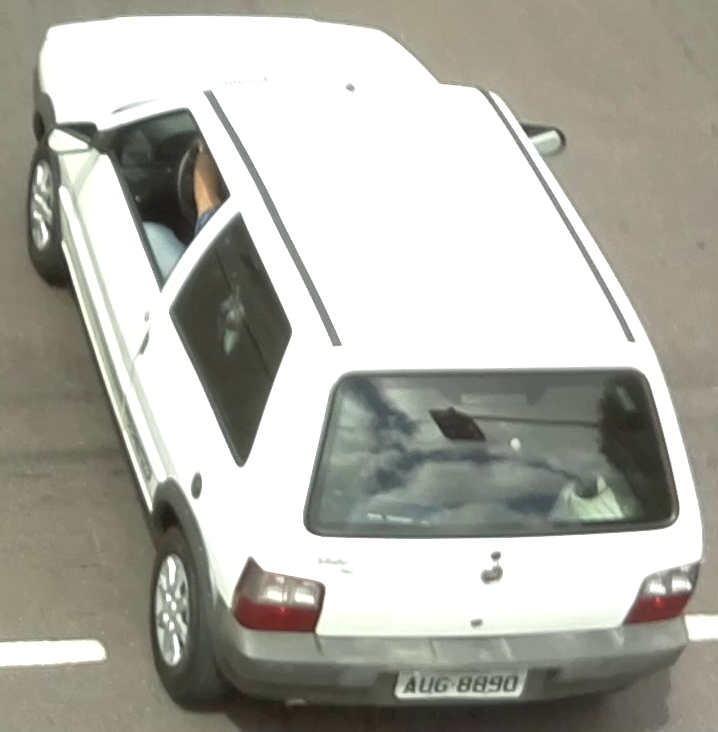}
};

\draw(5.0,7.00) node[] (car2) {
 \frame{\includegraphics[width=2.0cm,height=1.0cm]{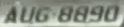}}
}; 
\end{tikzpicture}

\begin{tikzpicture}[scale=0.8, every node/.style={scale=0.8}]

\draw(0.0,6.0) node[] (car1) {
\includegraphics[width=3.8cm,height=3.2cm]{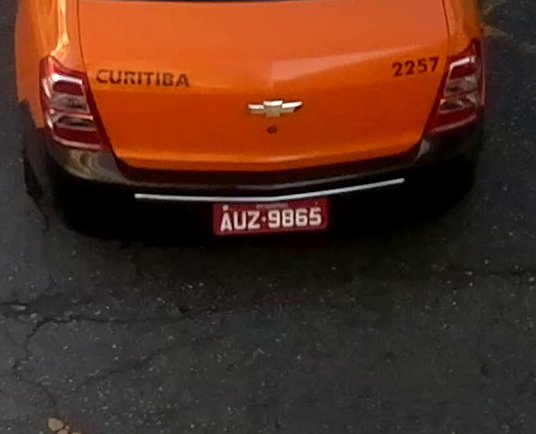}
};

\draw(2.1,4.13) node[text centered, text width=6.8cm] {
\footnotesize (c)
};

\draw(0.8,7.00) node[] (car1) {
 \frame{\includegraphics[width=2.0cm,height=1.0cm]{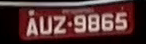}}
};

\draw(4.2,6.0) node[] (car2) {
\includegraphics[width=3.8cm,height=3.2cm]{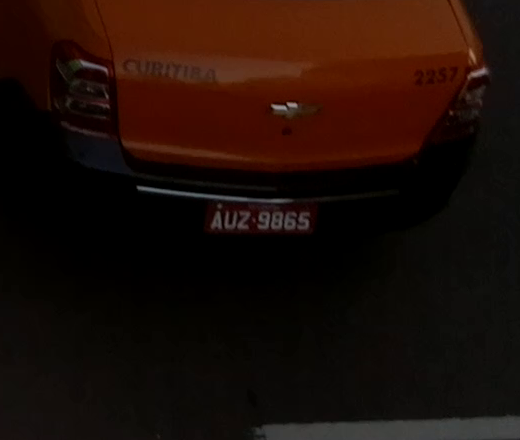}
};

\draw(5.0,7.00) node[] (car2) {
 \frame{\includegraphics[width=2.0cm,height=1.0cm]{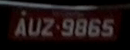}}
}; 
\end{tikzpicture}

\vspace{-2mm}
\caption{\small Examples of challenging scenarios for vehicle identification: \minor{(a)~similar vehicles with different license plates; (b)~similar license plate strings and distinct vehicles; and (c)~same vehicle under different lighting conditions.
The combination of attributes, e.g. vehicle appearance and textual information from the license plate region, can help to improve the recognition since two similar vehicles may have considerably different license plates and vice~versa.}
}
\label{fig:scenarios}
\end{figure}
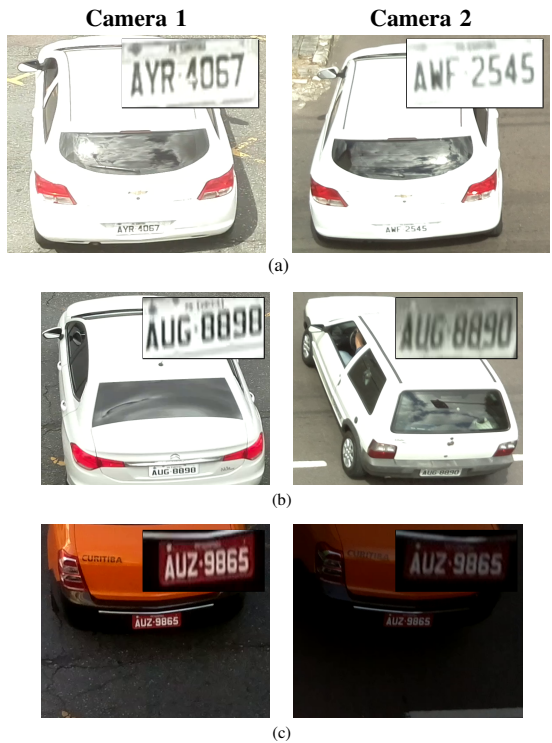

Although extensively investigated~\cite{8692748,5763781,5659904,8265213,8036238,tang2017multi}, this research problem is far from being solved since several challenges come from the high inter-class similarity, caused by vehicles of the same make, model and/or color that often look exactly the same, see Figure~\ref{fig:scenarios}(a), vehicles with similar license plate identifiers, see Figure~\ref{fig:scenarios}(b), and from the high intra-class dissimilarity, caused by abrupt illumination changes or camera viewpoints, that makes two instances of the same vehicle have differences, see Figure~\ref{fig:scenarios}(c). In the remainder of this section, we detail our research problem and the main contributions of this~work.

\subsection{Research problem}

\minor{The main issue of existing datasets for vehicle identification}~\cite{lou2019large, naphade2021ai_city_challenge, liu2016deep_relative} \minor{is the fact that the authors intentionally redacted the license plate identifier in all images to respect privacy restrictions, and, as explained later, the knowledge extracted from this unique identifier is essential for solving certain difficult matching problems, e.g., the correct identification of distinct but visually similar vehicles, as shown in Figure}~\ref{fig:scenarios}(a). 
\minor{However, in some regions/countries, the license plates are linked/related to the vehicle and not to the respective drivers/owners; in other words, in such cases it is not possible to obtain any public information about the vehicle owner based on the license plate.
One of the countries where this occurs is Brazil~}\cite{placa_veiculo_planalto}\minor{, where we collected images to create a novel dataset for vehicle identification that contains labeled license plate~information.}

\begin{figure}
  \centering
  \begin{tikzpicture}[scale=0.875, every node/.style={scale=0.875}]
   
     \draw(0.0,0.0) node[] (image) {
      \frame{\includegraphics[width=6.8cm, height=7.2cm]{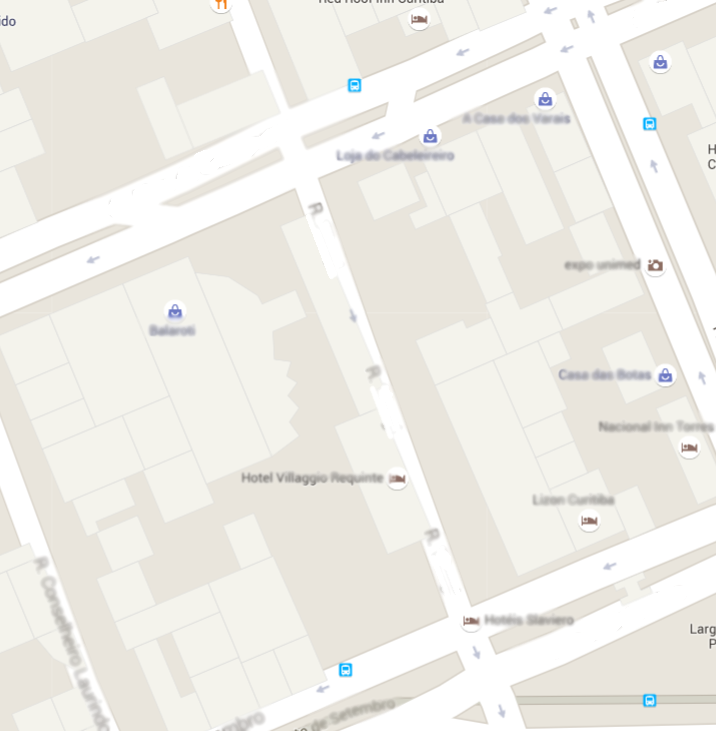}}   
     };
   
     \draw [line width=0.5mm, decoration={markings,mark=at position 1 with {\arrow[scale=1.5,>=stealth]{>}}},postaction={decorate}] (-1.0,2.5) -- (+0.81, -2.2);
     
     \draw [line width=0.5mm, decoration={markings,mark=at position 1 with {\arrow[scale=1.5,>=stealth]{>}}},postaction={decorate}] (-0.5,1.95) -- (+1.07, -2.1);
      
     \draw [line width=0.5mm] (-0.53,1.95) -- (+1.17, 2.67);
     
     \draw [line width=0.5mm, decoration={markings,mark=at position 1 with {\arrow[scale=1.5,>=stealth]{>}}},postaction={decorate}] (-0.75,1.85) -- (-2.50,1.05);
        
     \path[->](-1.5,3.2) node[] (s2) {\textbf{Camera 1}};
     \draw(-1.4,2.6) node[] (image) {
      \includegraphics[width=1.0cm]{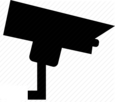} 
     };
     
     \path[->](+0.6,-3.0) node[] (s2) {\textbf{Camera 2}};
     \draw(0.5,-2.5) node[] (image) {
      \includegraphics[width=1.0cm]{./figures/curitiba/camera.png} 
     };
     
     \node[draw, fill=red, circle, scale=0.4](c1) at (-1.0,2.5) {};
     
     \node[draw, fill=red, circle, scale=0.4](c2) at (+1.0,-2.35) {};
     
      \node[draw, fill=black,circle, scale=0.52, text=white, inner sep=2pt](path1) at (-1.55,1.52) {\huge \textbf{B}};
      
      \node[draw, fill=black,circle, scale=0.52, text=white, inner sep=1pt](path1) at (-0.1,0.15) {\huge \textbf{A}};
      
      \node[draw, fill=black,circle, scale=0.52, text=white, inner sep=1pt](path1) at (-0.2,1.3) {\huge \textbf{C}};
     
     \draw(2.9,2.3) node[text centered, text width=2.7cm, inner sep=0pt] (image) { \frame{\includegraphics[width=2.7cm]{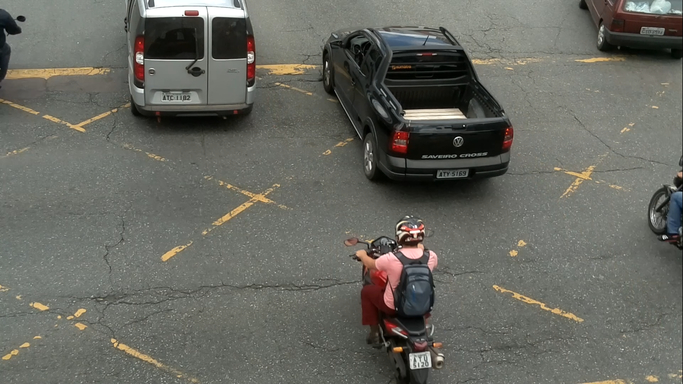}} };
     
     \draw(2.9,-2.3) node[text centered, text width=2.7cm, inner sep=0pt] (image) { \frame{\includegraphics[width=2.7cm]{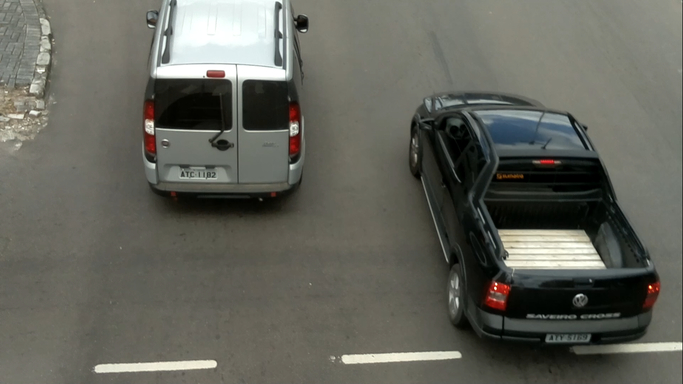}} };
     
  \end{tikzpicture}
  \caption{\small \minor{Illustration of the experimental environment setup}: a pair of low-cost Full-HD cameras, depicted by red dots, properly calibrated and time synchronized are monitoring two distinct traffic lights on the same street, 546 ft away. The road network is structured in such a way that some vehicles are monitored only by Camera~1, see route~B; only by Camera~2, see route~C; or by both cameras, see route~A.}  \label{fig:setup}
\end{figure}

\minor{In this work, we consider a road network topology structured as shown in  Figure}~\ref{fig:setup}, \minor{where the rear license plate is legible in most cases -- it is worth noting that in some countries/regions, e.g. several states in the United States, the license plate is attached only to the vehicle's rear.}
The images are taken from an elevated surveillance camera that records simultaneously multiple road lanes. Each vehicle of interest typically enters the field of view through the bottom part of the frame and leaves through the top side. As can be noted, not every vehicle seen in one camera appears in the other.

\subsection{Contributions}

This work has \minor{two} main contributions for the vehicle identification problem: 
\begin{itemize}
   \item \major{We introduce a novel dataset, called Vehicle-Rear, composed of high-resolution videos,
   that, to the best of our knowledge, is the first to consider the same camera view of most city systems used to enforce speed limits --~i.e., rear view of the vehicles with their license plates legible in most cases; Vehicle-Rear is associated with accurate information about each vehicle: make, model, color and~year, as well as the image coordinates of each license plate region and its corresponding ASCII string;}
   
   \item \major{We propose a novel two-stream CNN architecture that uses the most distinctive and persistent features for vehicle identification: coarse-resolution image patches, containing the \textit{vehicle shape}, feed one stream, while high-resolution \textit{license plate patches}, with string identifiers easily readable by humans (as present in the Vehicle-Rear dataset), feed the other stream.
   Such multi-resolution strategy helps to minimize the computational effort while it makes possible to capture the essential details for vehicle~identification;}
\end{itemize}

We believe that the creation of a publicly available dataset containing images captured in real-world scenarios and labeled information about both the vehicle and its license plate represents a step forward in designing different approaches to vehicle identification, since state-of-the-art algorithms for vehicle identification take advantage of only one of these attributes~\cite{8692748,8325486,deng2021trends}. We hope that our dataset and deep architecture can also be useful for other machine learning problems such as vehicle model identification, time-travel estimation, among~others.

\vspace{10pt}

The remainder of this paper is organized as follows. In Section~\ref{sec:related}, we review the literature on vehicle identification.
The proposed \dataset dataset is described in Section~\ref{sec:dataset}.
The two-stream architecture is described in Section~\ref{sec:method}, and the experimental evaluation is reported in Section~\ref{sec:experiments}.
In Section~\ref{sec:discussion} we discuss some alternative architectures, and in Section~\ref{sec:conclusions} we state the conclusions.

\section{Related work}~\label{sec:related}

Vehicle identification is an active field of research with many algorithms and an extensive bibliography.
As observed by Tian~{\etal}~\cite{6875912}, this problem is still an open issue for future developments of networked video surveillance systems, in which the road camera infrastructure is used to extract vehicle trajectories for behavior analysis and pattern discovery. Traditionally, algorithms proposed for this task were based on the comparison of electromagnetic signatures captured from a pair of inductive or magnetic sensors \cite{5983819,6229117}. This class of systems can benefit from the existing infrastructure to capture vehicle signature profiles from inductive-loop detectors~\cite{5763781}, weight-in-motion devices~\cite{Christiansen1996}, and microloop sensors~\cite{5659904}. However, as stated by Ndoye~\etal~\cite{5659904}, such signature-based algorithms are complex and depend on complicated data models or extensive~calibrations.

Video-based algorithms have been proven essential for vehicle identification.
As describe in the surveys of \minor{Deng~\textit{et al.}}~\cite{deng2021trends}\minor{,  Wang~\textit{et al.}}~\cite{wang2019survey}\minor{, and Khan \& Ullah}~\cite{khan2019survey}\major{, handcrafted image descriptors}~\cite{cabrera2011efficient,lowe2004sift,lbp,MINETTO2013} \major{were the first attempt to solve this problem, e.g. Zapletal and
Herout}~\cite{zapletal2016vehicle} and
\major{Chen~\textit{et al.}}~\cite{chen2018real} used HOG descriptors, \minor{Cabrera ~\textit{et al.}}~\cite{cabrera2011efficient}  \major{ used HAAR descriptors}, while \minor{Cormier~\textit{et al.}} ~\cite{7470114} \major{used Local Binary Patterns (LBP)}~\cite{lbp}
\major{-- all these works also combined other hand-crafted descriptors}.
\minor{Zhang~\textit{et al.}}~\cite{Zhang2016} used Scale-Invariant Feature Transform~(SIFT)~\cite{lowe2004sift} \major{to distinguish between subordinate categories with similar visual appearance, caused by a huge number of car design and models with similar appearance.
In particular, SIFT was widely explored to extract distinctive key points from the vehicle for feature correspondence}~\cite{Luvizon:2016}.

The use of Siamese-based architectures for the specific problem of vehicle identification is common.
Tang~\etal~\cite{tang2017multi} proposed to fuse deep and handcrafted features using a \emph{Triplet Siamese  Network}~\cite{10.1007/978-3-319-24261-3_7} -- a network that attempts to minimize the distance between an anchor and a positive sample and to maximize the distance between the same anchor and a negative sample. Yan~\etal~\cite{8265213} proposed a novel Triplet Loss Function, which uses both the intra-class variance and the inter-class similarity in vehicle models, but using only vehicle shape features.
Liu~\etal~\cite{8036238} developed a coarse-to-fine algorithm for vehicle identification that filters out potential matchings with handcrafted and deep features based on color and shape, and then used a Siamese network for the license plate regions. 

The idea of multi-stream \glspl*{cnn} has also been considered by many authors to tackle different identification problems. Ye~\etal~\cite{Ye:2015:ETC:2671188.2749406} proposed a two-stream architecture that uses static video frames and optical flow features for video classification. Similarly, Chung~\etal~\cite{chung2017two} proposed a two-stream Siamese architecture that is also based on spatial and temporal information extracted from RGB frames and optical flow features but for person re-identification. Zagoruyko~\etal~\cite{zagoruyko2015} described distinct Siamese architectures to compare image patches. 
In particular, they developed a two-stream architecture that explores multi-resolution information by using the central part of an image patch and the surrounding part of the same patch.
Specifically for vehicle identification, Oliveira~\etal~\cite{icaro2019} proposed a two-stream network fed by small patches from the vehicle shape and the license plate region, and Guo~\etal~\cite{8692748} proposed a three-stream network where one stream extracts global features from the vehicle shape and the other two streams learn to locate vehicle features, such as windscreen and car-head parts.

Architectures designed to recognize patterns in temporal sequences, such as \gls*{lstm}~\cite{hochreiter1997lstm}, ensembles~\cite{8698456}, and spatio-temporal~(3D) convolutions~\cite{6165309}, may also have a major impact on vehicle identification~\cite{Shen_2017_ICCV,8354181}. As an example, Shen~\etal~\cite{Shen_2017_ICCV} noted that if a vehicle is seen by cameras 1 and 3 then it should also appear in camera 2; thus, if no candidate is observed by camera 2, any subsequent match should have very low confidence. 
The authors employed a Siamese network fed with the vehicle's shape and temporal metadata to model this scenario,
and an \gls*{lstm} to evaluate the visual and spatio-temporal differences of neighboring states along with path proposals. The dataset used in their experiments, \verisete~\cite{liu2016deep_learning}, was acquired by 20 cameras. Zhou~\etal~\cite{8354181} exploited an adversarial bi-directional \gls*{lstm} network to create a vehicle representation from one camera view that would allow modeling transformations across continuous view~variations.
\glspl*{gan} were also explored to generate samples to facilitate the vehicle identification task~\cite{8653852}.

License plate recognition, as we used in this work, is one of the key attributes for successful vehicle identification and deep networks have achieved many advances in this field.
Li~\etal~\cite{li2018reading} first extracted sequential features from the license plate patch using a \gls*{cnn} in a sliding window manner. 
Then, \glspl*{brnn} with \gls*{lstm} were applied to label the sequential features, while \gls*{ctc} was employed for sequence decoding. The results showed that their method attained better recognition rates than the baselines. Nevertheless, Dong~\etal~\cite{dong2017cnnbased} claimed that such a method is very fragile to distortions caused by viewpoint change and therefore is not suitable for license plate recognition in the wild. Thus, a license plate rectification step is employed first in their approach, which leverages parallel \glspl*{stn} with shared-weight classifiers.
Recently, Selmi et al.~\cite{selmi2020delpdar} trained a Mask-RCNN~\cite{he2017maskrcnn} to predict $37$ positive classes ($0$-$9$, A-Z, and one Arabic word).
Despite the fact that promising results were reported in their experiments, the chosen model (with an input size of $530\times300$ pixels) is much more computationally expensive than those used in other works (e.g.,~\cite{goncalves2018realtime,silva2020realtime,laroca2021efficient}) for license plate recognition, which makes it difficult (or even impossible) for it to be employed in some real-world applications --~especially those where multiple vehicles can coexist on the~scene.

Silva \& Jung~\cite{silva2017realtime} proposed a YOLO-based model to simultaneously detect and recognize all characters within a cropped license plate.
While impressive \gls*{fps} rates were reported in their experiments, less than 65\% of the license plates on the test set were correctly recognized since the character classes in the training set used by them were highly unbalanced.
Accordingly, Laroca \etal~\cite{laroca2018robust,laroca2021efficient} and Silva \& Jung~\cite{silva2018license,silva2020realtime} retrained that model, called CR-NET, with enlarged training sets composed of real images and many other artificially generated.
In all these works, the retrained networks became much more robust for the detection and classification of real~characters.

\minor{As final remarks, although some previous studies have shown the importance of feature fusion for vehicle identification (e.g.,~}\cite{8036238, tang2017multi,8692748}\minor{), none of them explored a camera infrastructure specifically designed for traffic law enforcement as those available in many cities, where the vehicle's rear license plate is legible in most cases.}
Considering such camera views, it is possible to develop a novel and robust two-stream architecture that combines two decisive features for vehicle identification: (i)~shape features from the vehicle rear-end and (ii)~textual features from the license plate~region. 
\section{The Vehicle-Rear Dataset} 
\label{sec:dataset}

As detailed in Table~\ref{tab:dataset}, the \dataset dataset consists of $10$ videos -- five from Camera~$1$ and five from Camera~$2$ (20 minutes long each video) -- captured by a low-cost 5-megapixel CMOS image sensor, time-synchronized, with a resolution of $1{,}920\times1{,}080$ pixels at 30.15 frames per second.

\begin{table}[!htb]
   \setlength{\tabcolsep}{4pt}
   \def\mc{\multicolumn{2}{c}}
   \renewcommand{\arraystretch}{1.1}
   \centering
   \caption{\small 
   \minor{Vehicle-Rear dataset}: detailed information about the number of vehicles, with and without a legible license plate, recorded by Cameras~$1$ and~$2$; and the number of true matchings between Camera~$1$ and~$2$.
   }
   \vspace{1mm}
   \resizebox{0.99\linewidth}{!}{
    \begin{tabular}{cccccc}
         \toprule
                &  \mc{Camera $1$} & \mc{Camera $2$} &   \\ \midrule
                  Set &  \# Vehicles &  \# Plates &  \# Vehicles &  \# Plates & \# Matchings \\ \midrule
           $01$ & $385$ & $342$ & $277$ & $245$ & $199$   \\
           $02$ & $350$ & $301$ & $244$ & $225$ & $179$   \\
           $03$ & $340$ & $312$ & $273$ & $252$ & $203$   \\
           $04$ & $280$ & $258$ & $230$ & $196$ & $147$   \\
           $05$ & $345$ & $299$ & $242$ & $205$ & $165$   \\ \midrule
        Total & $1{,}700$ & $1{,}512$ & $1{,}266$ & $1{,}123$ & $893$  \\ \bottomrule
   \end{tabular}
   }
   \label{tab:dataset}
 \end{table}
 
\begin{figure}[!htb]
    \setlength{\tabcolsep}{3pt}
    \def\mc{\multicolumn{3}{c}}
    \renewcommand{\arraystretch}{1.0}
    \begin{tabular}{ccc}

    \mc{\cellcolor{blue!40!gray!30} \footnotesize Camera $1$}\\
    \cellcolor{blue!40!gray!30} \includegraphics[scale=0.15]{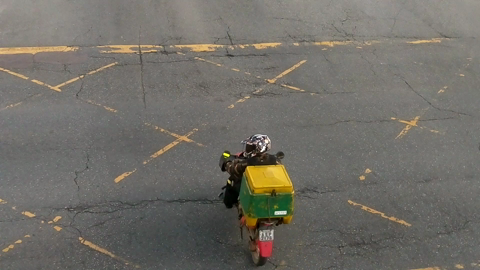} & \cellcolor{blue!40!gray!30} \includegraphics[scale=0.15]{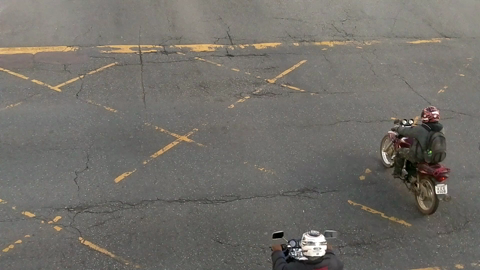} 
    & \cellcolor{blue!40!gray!30} \includegraphics[scale=0.15]{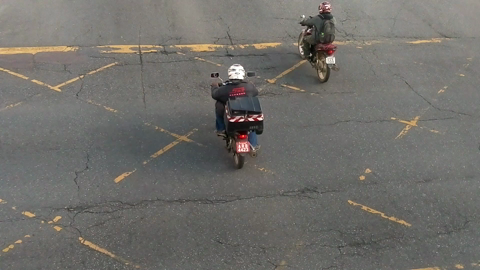}
    \\ 
    \mc{\cellcolor{green!20!gray!30} \footnotesize Camera $2$}\\
    \cellcolor{green!20!gray!30}
    \includegraphics[scale=0.15]{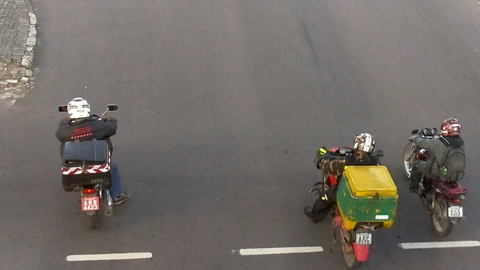} & \cellcolor{green!20!gray!30}
    \includegraphics[scale=0.15]{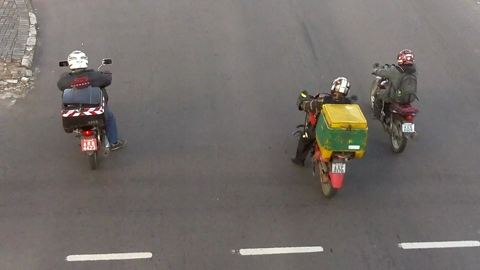} 
    & \cellcolor{green!20!gray!30}
    \includegraphics[scale=0.15]{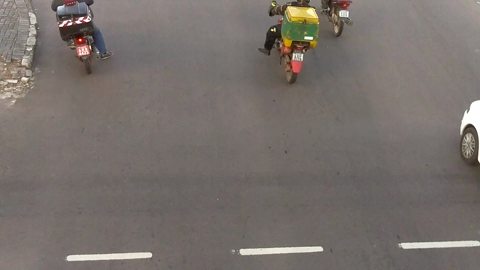}\\
    \mc{\footnotesize \minor{(a)}} \\[1ex] 
    
    \mc{\cellcolor{blue!40!gray!30} \footnotesize Camera $1$}\\
    \cellcolor{blue!40!gray!30} \includegraphics[scale=0.15]{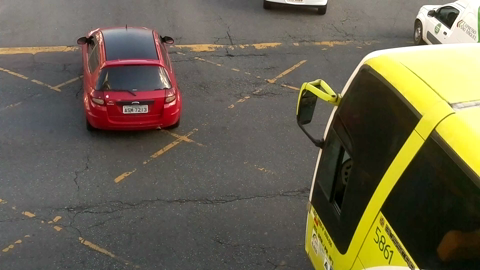} & \cellcolor{blue!40!gray!30} \includegraphics[scale=0.15]{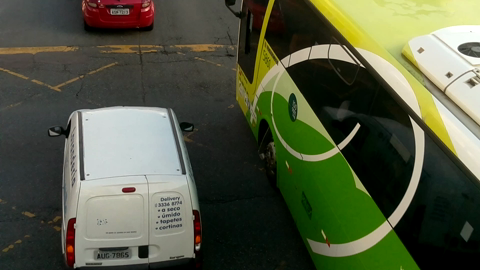} 
    & \cellcolor{blue!40!gray!30} \includegraphics[scale=0.15]{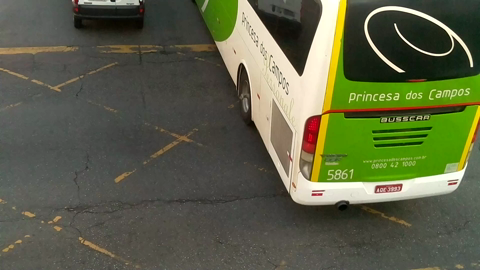}\\
    
    \mc{\cellcolor{green!20!gray!30} \footnotesize Camera $2$}\\
    \cellcolor{green!20!gray!30}
    \includegraphics[scale=0.15]{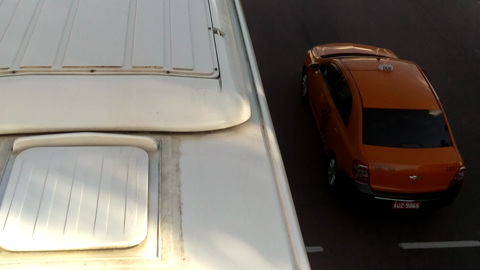} & \cellcolor{green!20!gray!30}
    \includegraphics[scale=0.15]{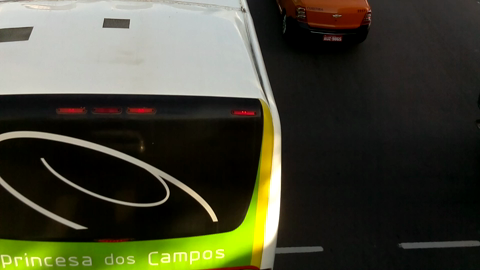} 
    & \cellcolor{green!20!gray!30}
    \includegraphics[scale=0.15]{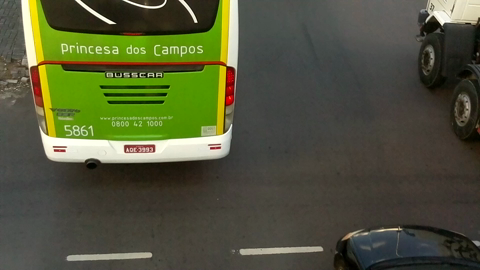}\\
    \mc{\footnotesize \minor{(b)}}\\[1ex]
    
    \mc{\cellcolor{blue!40!gray!30} \footnotesize Camera $1$}\\
    \cellcolor{blue!40!gray!30} \includegraphics[scale=0.15]{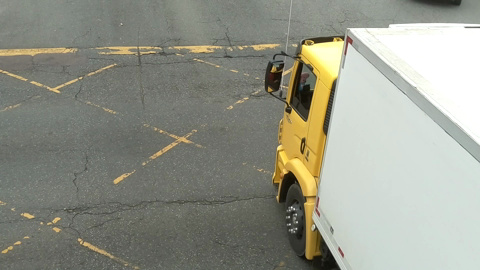} & \cellcolor{blue!40!gray!30} \includegraphics[scale=0.15]{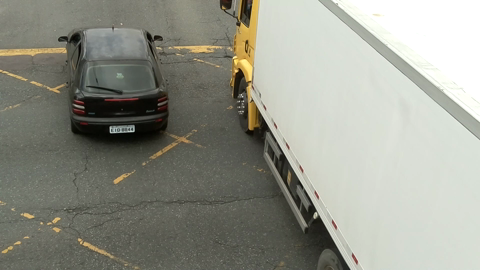} 
    & \cellcolor{blue!40!gray!30} \includegraphics[scale=0.15]{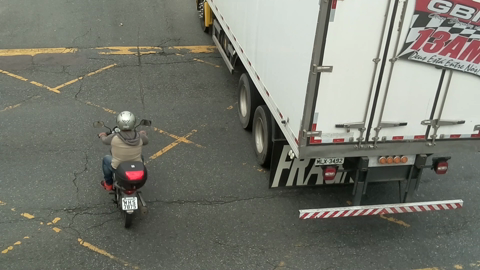}\\
    \mc{\cellcolor{green!20!gray!30} \footnotesize Camera $2$}\\
    \cellcolor{green!20!gray!30} \includegraphics[scale=0.15]{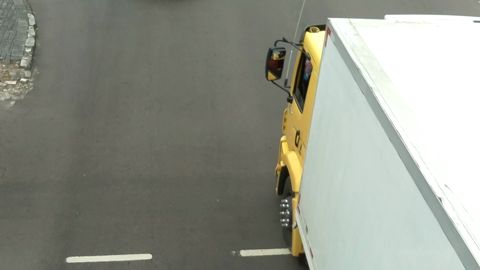} & \cellcolor{green!20!gray!30} \includegraphics[scale=0.15]{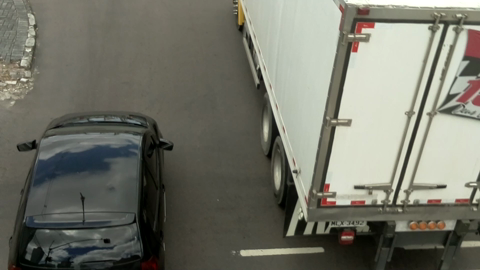} 
    & \cellcolor{green!20!gray!30} \includegraphics[scale=0.15]{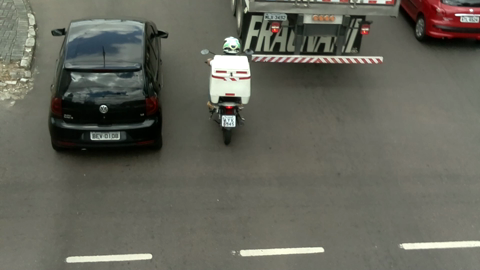}\\
    \mc{\footnotesize \minor{(c)}}\\[1ex]

    \mc{\cellcolor{blue!40!gray!30} \footnotesize Camera $1$}\\
    \cellcolor{blue!40!gray!30} \includegraphics[scale=0.15]{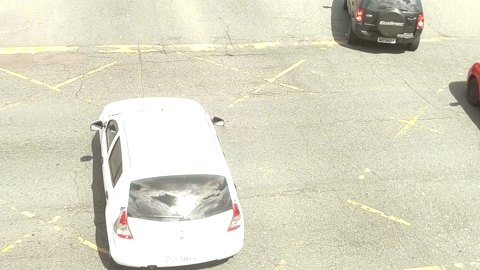} & \cellcolor{blue!40!gray!30} \includegraphics[scale=0.15]{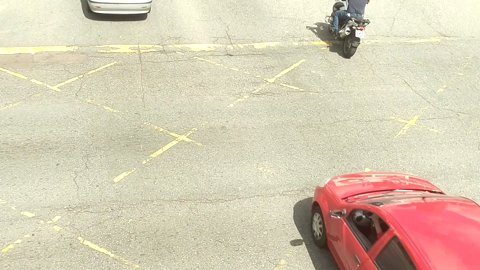} 
    & \cellcolor{blue!40!gray!30} \includegraphics[scale=0.15]{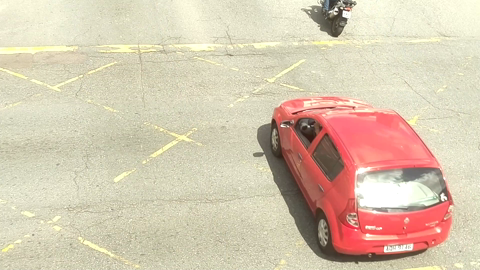}\\
    \mc{\cellcolor{green!20!gray!30} \footnotesize Camera $2$}\\
    \cellcolor{green!20!gray!30} \includegraphics[scale=0.15]{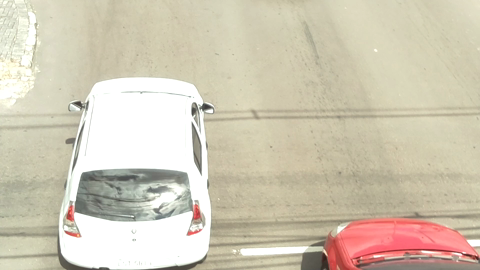} & \cellcolor{green!20!gray!30} \includegraphics[scale=0.15]{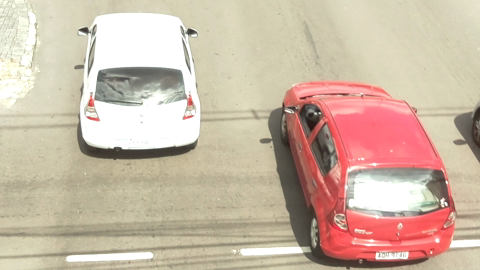} 
    & \cellcolor{green!20!gray!30} \includegraphics[scale=0.15]{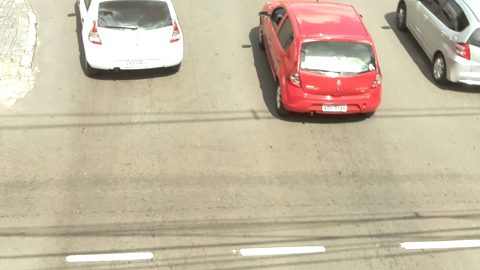}\\
    \mc{\footnotesize \minor{(d)}}\\
    \end{tabular}
    \caption{\small Image sequences from the proposed Vehicle-Rear dataset. \minor{The temporal sequences show examples of (a)~motorcycles; (b)~cars and buses; (c)~trucks; (a) and~(c)~in normal weather conditions; (b)~dark frames caused by the motion of large vehicles; and (d)~severe lighting conditions.}}
    \label{fig:lighting_conditions}
\end{figure}

We chose a busy avenue of the city, with traffic of different types of vehicles, and different periods of the day to record the videos so that each set has very specific lighting conditions (see Figure~\ref{fig:lighting_conditions}). 
Note that temporal information can also be explored in the \dataset dataset since for each vehicle we have between [$5$-$25$] frame occurrences per camera (depending on the vehicle speed); thus, redundant information could be used to further improve the vehicle~identification.

For each video, we provide a ground truth XML file in which each entry, corresponding to a distinct vehicle, has an axis-aligned rectangular box of the first license plate occurrence, the corresponding identifier in ASCII code, the frame position, as well as the vehicle's make, model, color and year, which were recovered from the database of the \gls*{denatran}. \minor{We remark that the DENATRAN database is publicly available, that is, there is no restriction on access to such information}. As far as we are aware, the proposed dataset is the first public dataset for vehicle identification to provide information on the appearance of the vehicles and also on their license~plates.

\begin{figure}[!htb]
    \centering
    \includegraphics[width=0.95\linewidth]{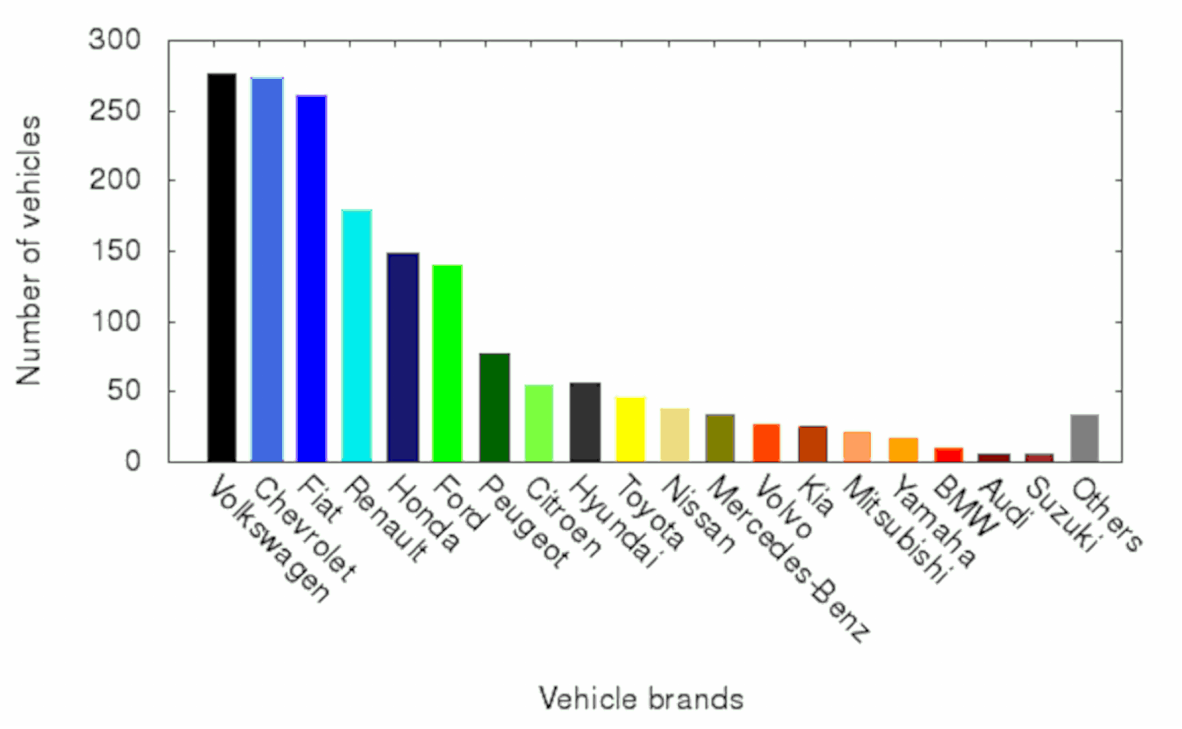}
    
    \vspace{-2mm}
    \caption{\small Vehicle \minor{histogram} by brand in the \dataset dataset.}
    \label{fig:hist_fine_grained1}
\end{figure}

\begin{figure}[!htb]
    \centering
    \includegraphics[width=0.95\linewidth]{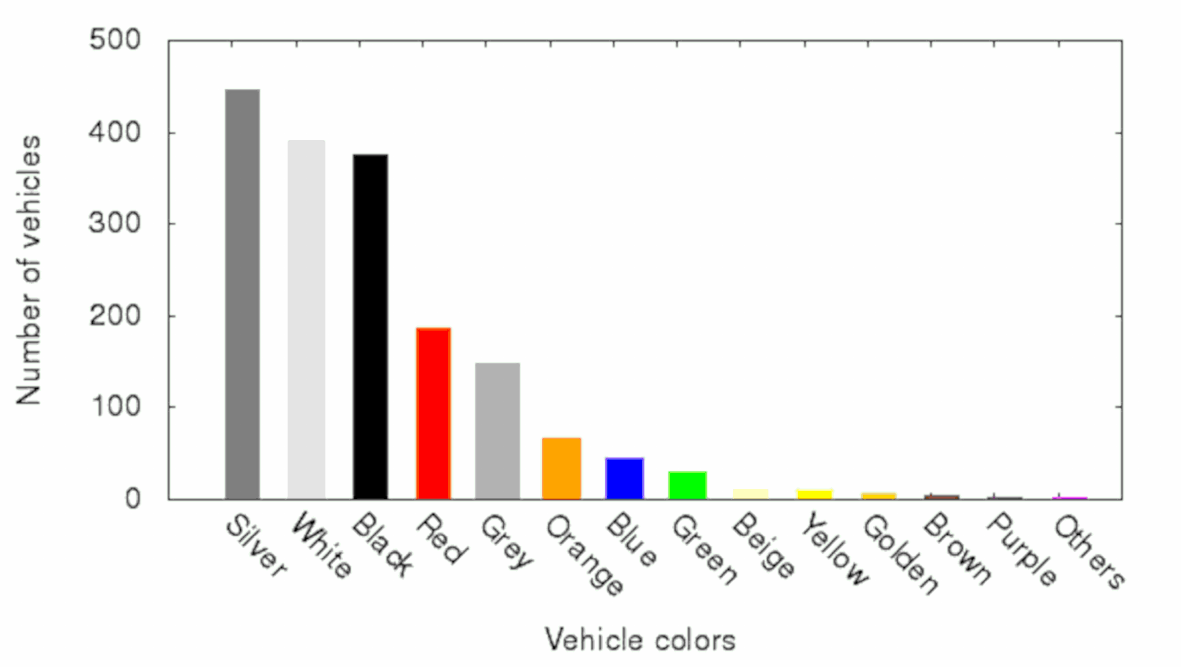}
    \vspace{-2mm}
    \caption{\small Vehicle \minor{histogram} by color in the \dataset dataset.}
    \label{fig:hist_fine_grained2}
\end{figure}

Figure~\ref{fig:hist_fine_grained1} and Figure~\ref{fig:hist_fine_grained2} show the diversity of our dataset in relation to vehicle automakers and colors,~respectively.
\minor{As can be seen, there is a considerable imbalance --~as is likely the case for every dataset~-- since vehicles of certain brands and colors sell more than others}.
Nevertheless, according to our experiments, such imbalance did not significantly affect the results obtained by the evaluated~models.

\minor{Finally, it is worth noting that the licenses plates of vehicles in Brazil, where the images were collected, are linked/related to the vehicle and no public information is available about the vehicle drivers/owners;~} \major{hence, a license plate remains the same after a change in vehicle ownership~} \cite{placa_wikipedia}\major{. Considering the height and distance of the cameras, as well as the fact that they record the rear view of vehicles, identifying the driver/owner from the captured frames in our dataset is not possible, to the best of our knowledge. 
Finally, as detailed in Section~}\ref{sec:discussion}\major{, this study was officially authorized to collect and explore open data such as the Vehicle-Rear~dataset.}

\section{Vehicle Identification Architecture}
\label{sec:method}

In order to explore the attributes of the proposed dataset, we design a two-stream neural network, as shown in Figure~\ref{fig:system-overview}, that uses the most distinctive and persistent features available for vehicle identification: coarse-resolution image patches, containing the vehicle shape, feed one stream, while high-resolution license plate patches, easily readable by humans, feed the other stream.
Such a multi-resolution strategy helps to minimize the computational effort while making it possible to capture the necessary details for the recognition. 
\minor{We developed a text descriptor, i.e., } \gls*{ocr}, which is combined with the shape descriptor through a sequence of fully connected layers for decision. Further details on these key steps are presented in the remainder of this~section.

\input{figures/two_stream.tex}

\subsection{Preliminaries}\label{sec:preliminaries}

\begin{figure}[!htb]
 \centering
 \begin{tikzpicture}[ele/.style={fill=black,circle,minimum width=.6pt,inner sep=1pt},every fit/.style={ellipse,draw,inner sep=-2pt}]
 
  \path[->](-0.2,4.2) node[] (s1) { $\mathbf{\mathcal{C}}$ };
  \path[->](+3.4,4.2) node[] (s1) { $\mathbf{\mathcal{N}}$ };
  \path[->](+1.6,3.8) node[] (s1) { $\mathbf{f}$ };

  \footnotesize
  \node[ele,label=left:$A$] (a1) at (0,3.5) {};    
  \node[ele,label=left:$B$] (a2) at (0,3.2) {};    
  \node[] (a4) at (-0.2,2.9) {$\dots$};
  \node[ele,label=left:$Z$] (a5) at (0,2.6) {}; 
  \node[ele,label=left:$0$] (a6) at (0,2.3) {};
  \node[ele,label=left:$1$] (a7) at (0,2.0) {};
  \node[] (a8) at (-0.2,1.7) {$\dots$};
  \node[ele,label=left:$9$] (a9) at (0,1.4) {};
  
  \node[ele,label=right:$0.00$] (b1) at (3.0,3.5) {};    
  \node[ele,label=right:$0.02$..] (b2) at (3.0,3.2) {};    
  \node[] (b4) at (3.4,2.9) {$\dots$};
  \node[ele,label=right:$0.71$..] (b5) at (3.0,2.6) {}; 
  \node[ele,label=right:$0.74$..] (b6) at (3.0,2.3) {};
  \node[ele,label=right:$0.77$..] (b7) at (3.0,2.0) {};
  \node[] (b8) at (3.4,1.7) {$\dots$};
  \node[ele,label=right:$1.00$] (b9) at (3.0,1.4) {};

  \node[draw,fit= (a1) (a2) (a4) (a5) (a6) (a7) (a8) (a9),minimum width=2.0cm] {} ;
  \node[draw,fit= (b1) (b2) (b4) (b5) (b6) (b7) (b8) (b9),minimum width=2.0cm] {} ;  
  \draw[->,shorten <=2pt,shorten >=2pt] (a1) -- (b1);
  \draw[->,shorten <=2pt,shorten >=2] (a2) -- (b2);
  \draw[->,shorten <=2pt,shorten >=2] (a5) -- (b5);
  \draw[->,shorten <=2pt,shorten >=2] (a6) -- (b6);
  \draw[->,shorten <=2pt,shorten >=2] (a7) -- (b7);
  \draw[->,shorten <=2pt,shorten >=2] (a9) -- (b9);
  
  \node[] (a8) at (1.6,2.9) {$\dots$};  
  \node[] (a8) at (1.6,1.7) {$\dots$};
 \end{tikzpicture}
 \caption{A bijective function ($f$) to map license plate characters (domain $\mathcal{C}$) to real numbers (range $\mathcal{N}$).}
 \label{fig:bijection}
\end{figure}
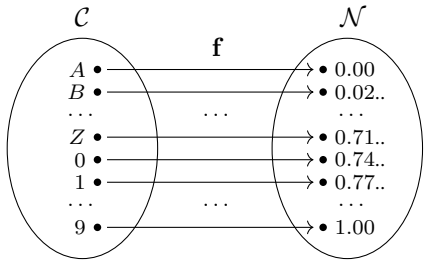

For our problem, let $S^{(c_1)} = \langle s^{(c_1)}_1, s^{(c_1)}_2, \dots, s^{(c_1)}_m \rangle$ and
$S^{(c_2)} = \langle s^{(c_2)}_1, s^{(c_2)}_2, \dots, s^{(c_2)}_m \rangle$ be two $m$-dimensional vectors representing the deep features extracted with a Siamese network from shape patches recorded by cameras $1$ and $2$, respectively. Also, let $\mathcal{C}_{n} = \{c_0, c_1, \dots, c_{n-1}\}$ be a non-empty alphabet consisting of $n$ unique elements. Then, let $f: \mathcal{C} \rightarrow \mathcal{N}$ be a one-to-one function (bijection) that maps
elements of the alphabet $\mathcal{C}$ to unique real numbers $\mathcal{N}$ \minor{according to Equation}~(\ref{eq:mapping}) 
\begin{equation}
  f(c_i) =  \frac{i}{n-1}
\label{eq:mapping}
\end{equation}
where $i$ is the element position in the alphabet, such that $0 \leq i < n$, and $n$ denotes the set size. The alphabet used to build the license plate identifiers is composed by 26~letters and 10~digits, thus, $\mathcal{C}_{36} = \{A, \dots, Z, 0, \dots, 9\}$.
This mapping is shown in Figure~\ref{fig:bijection}.
Note that the lexicography order is used to establish the mapping function $f$.
As a consequence, no special arrangement among similar characters, such as $D$, $O$, $Q$ and $0$, was done.

\subsection{Shape similarities}

The shape similarities are identified by a Siamese network, which hereinafter is referred
to as Shape-Stream.
This particular class of neural architecture was introduced by Bromley \etal~\cite{Bromley:1993:SVU:2987189.2987282} and consists of two identical networks that share the same weights. \major{We choose a Siamese network to compare shape similarities because it is an effective and simple architecture to solve image matching problems.}

The shape descriptor is defined as a new vector \minor{according to Equation}~(\ref{eq:shape_descritor})
\begin{equation}
   S = S^{(c_1)} - S^{(c_2)} = (s_1, s_2, \dots, s_m)
   \label{eq:shape_descritor}
\end{equation}
where each component $s_i$ is given by an $L_1$ (Manhattan) distance, that is, $s_i = |s^{(c_1)}_i - s^{(c_2)}_i|$ for cameras $c_1$ and~$c_2$.
The twin networks guarantee that two similar image patches will not be mapped to very different locations in the feature space since they compute the same function and their weights are tied; therefore, it is expected that the vector components are small for two instances of the same vehicle and large otherwise.
The deep features were extracted with a low complex VGG-based \gls*{cnn}~\cite{icaro2019}, called Small-VGG, formed by a reduced number of convolutional layers in order to save computational effort, as shown in Table~\ref{tab:cnn_model}. 
\begin{table}[!htb]
\footnotesize
\centering
\caption{\small The \gls*{cnn} architecture used by the Siamese network in the Shape-Stream.}
\vspace{1mm}
\resizebox{0.99\linewidth}{!}{ 
\begin{tabular}{@{}cccccc@{}}
\toprule
\textbf{\#} & \textbf{Layer} & \textbf{Filters} & \textbf{Size} & \textbf{Input} & \textbf{Output} \\ \midrule
$0$ & conv & $64$ & $3 \times 3 / 1$ & $64 \times 64 \times 3$ & $64 \times 64 \times 64$ \\
$1$ & max &  & $2 \times 2 / 2$ & $64 \times 64 \times 64$ & $32 \times 32 \times 64$ \\
$2$ & conv & $128$ & $3 \times 3 / 1$ & $32 \times 32 \times 64$ & $32 \times 32 \times 128$\\
$3$ & max & & $2 \times 2 / 2$ & $32 \times 32 \times 128$ & $16 \times 16 \times 128$\\
$4$ & conv & $128$ & $3 \times 3 / 1$ & $16 \times 16 \times 128$ & $16 \times 16 \times 128$\\
$5$ & max & & $2 \times 2 / 2$ & $16 \times 16 \times 128$ & $8 \times 8 \times 128$\\
$6$ & conv & $256$ & $3 \times 3 / 1$ & $8 \times 8 \times 128$ & $8 \times 8 \times 256$\\
$7$ & max & & $2 \times 2 / 2$ & $8 \times 8 \times 256$ & $4 \times 4 \times 256$\\
$8$ & conv & $512$ & $3 \times 3 / 1$ & $4 \times 4 \times 256$ & $4 \times 4 \times 512$\\
$9$ & max & & $2 \times 2 / 2$ & $4 \times 4 \times 512$ & $2 \times 2 \times 512$\\ \bottomrule
\end{tabular} \,
}
\label{tab:cnn_model}
\end{table}

\subsection{License plate similarities}

The plate similarities are then identified by using textual information extracted from fine-resolution license plate image patches~(OCR-Stream).
We observed through a series of experiments, as detailed in Section~\ref{sec:discussion}, that the same approach we used for shape was not very accurate to distinguish between very similar license plate regions.
The textual content, on the other hand, makes it possible to explore the syntax that defines the license plate layouts and, thus, to improve the recognition.
Inspired by the tremendous advances in machine learning achieved by \glspl*{cnn}, we used a state-of-the-art architecture (CR-NET)~\cite{silva2017realtime} for OCR that has proven to be robust to recognize license plates from various countries~\cite{silva2020realtime,laroca2021efficient}, but here it was fine-tuned for the Brazilian license plate layout (i.e., three letters followed by four~digits). 

The OCR architecture, as described by
Silva \& Jung~\cite{silva2017realtime} and later improved 
by Laroca~\etal~\cite{laroca2021efficient}, consists of the first eleven layers of YOLO~\cite{redmon2016yolo} and four other convolutional layers added to improve non-linearity, as shown in Table~\ref{tab:cr_net}.
The network was trained to predict $35$ character classes ($0$-$9$, A-Z, where the letter~`O' is detected/recognized jointly with the digit~`0') -- however, for the sake of simplicity of definitions, we will assume a complete alphabet with $36$~characters in the remainder of this section. Furthermore, some swaps of digits and letters, which are often misidentified, were used to improve the recognition: [$1$~$\Rightarrow$~I; $2$~$\Rightarrow$~Z; $4$~$\Rightarrow$~A; $5$~$\Rightarrow$~S; $6$~$\Rightarrow$~G; $7$~$\Rightarrow$~Z; $8$~$\Rightarrow$~B] and [A~$\Rightarrow$~$4$; B~$\Rightarrow$~$8$; D~$\Rightarrow$~$0$; G~$\Rightarrow$~$6$; I~$\Rightarrow$~$1$; J~$\Rightarrow$~$1$; Q~$\Rightarrow$~$0$; S~$\Rightarrow$~$5$;~Z~$\Rightarrow$~$7$].

\begin{table}[!htb]
\centering
\caption{\small The CNN-OCR architecture for license plate recognition as proposed by Silva \& Jung~\cite{silva2017realtime} and improved 
by Laroca~\etal~\cite{laroca2021efficient}.}
\vspace{1mm}
\resizebox{0.99\linewidth}{!}{ 
\begin{tabular}{@{}cccccc@{}}
\toprule
\textbf{\#} & \textbf{Layer} & \textbf{Filters} & \textbf{Size} & \textbf{Input} & \textbf{Output} \\ \midrule
$0$ & conv & $32$ & $3 \times 3 / 1$ & $352 \times 128 \times 3$ & $352 \times 128 \times 32$ \\
$1$ & max &  & $2 \times 2 / 2$ & $352 \times 128 \times 32$ & $176 \times 64 \times 32$ \\
$2$ & conv & $64$ & $3 \times 3 / 1$ & $176 \times 64 \times 32$ & $176 \times 64 \times 64$\\
$3$ & max &  & $2 \times 2 / 2$ & $176 \times 64 \times 64$ & $88 \times 32 \times 64$ \\
$4$ & conv & $128$ & $3 \times 3 / 1$ & $88 \times 32 \times 64$ & $88 \times 32 \times 128$ \\
$5$ & conv & $64$ & $1 \times 1 / 1$ & $88 \times 32 \times 128$ & $88 \times 32 \times 64$ \\
$6$ & conv & $128$ & $3 \times 3 / 1$ & $88 \times 32 \times 64$ & $88 \times 32 \times 128$ \\
$7$ & max & & $2 \times 2 / 2$ & $88 \times 32 \times 128$ & $44 \times 16 \times 128$ \\
$8$ & conv & $256$ & $3 \times 3 / 1$ & $44 \times 16 \times 128$ & $44 \times 16 \times 256$ \\
$9$ & conv & $128$ & $1 \times 1 / 1$ & $44 \times 16 \times 256$ & $44 \times 16 \times 128$\\
$10$ & conv & $256$ & $3 \times 3 / 1$ & $44 \times 16 \times 128$ & $44 \times 16 \times 256$\\
$11$ & conv & $512$ & $3 \times 3 / 1$ & $44 \times 16 \times 256$ & $44 \times 16 \times 512$ \\
$12$ & conv & $256$ & $1 \times 1 / 1$ & $44 \times 16 \times 512$ & $44 \times 16 \times 256$ \\
$13$ & conv & $512$ & $3 \times 3 / 1$ & $44 \times 16 \times 256$ & $44 \times 16 \times 512$ \\
$14$ & conv & $200$ & $1 \times 1 / 1$ & $44 \times 16 \times 512$ & $44 \times 16 \times 200$ \\
$15$ & detection &  &  &  &  \\ \bottomrule
\end{tabular} \,
}
\label{tab:cr_net}
\end{table}

We created an \gls*{ocr} descriptor by combining the textual content extracted from both license plates. For that purpose, we propose a scheme to map  characters to real numbers as~follows. 

The OCR descriptor is composed by the mapped characters, alternated with its classification scores so as to aggregate knowledge about the confidence of each prediction.
Moreover, the descriptor also contains the similarities between both license plate identifiers. Namely, for two aligned strings, we compute a character-by-character distance using a step function, \minor{as shown in Equation}~(\ref{eq:distance})
\begin{equation}
d(c_i,c_j) = 
\begin{cases}
0 & \mbox{\textbf{if}} \;\; f(c_i) - f(c_j) = 0 \\ 
1 & \mbox{\textbf{otherwise}}
\end{cases}
\label{eq:distance}
\end{equation}
where $c_i$ and $c_j$ are two characters that belong to set $\mathcal{C}_{36}$, \minor{detailed in Section}~\ref{sec:preliminaries},  and $f$ is the mapping function of Equation~(\ref{eq:mapping}).
Observe that two characters are equal or distinct for the step function, i.e.,~the notion of proximity does not exist. For example, although
letter A is mapped to value 0.00, B to 0.02... and Z to 0.71..., the distance between A and B is the same distance between A and Z ($1$ for both cases).
However, the confidence scores, associated with each character, may help the network to decide the weight of such distances.

The OCR descriptor is illustrated in Figure~\ref{fig:ocr_descriptor}.
 
\input{figures/plate_descriptor.tex}
\section{Experiments}
\label{sec:experiments}

In this section, we describe an extensive set of experiments comparing several \gls*{cnn}/\gls*{ocr} architectures. 

\begin{figure}[!htb]
\centering
\subfigure[Non-matching pairs]{
\begin{tikzpicture}[scale=0.8, every node/.style={scale=0.8}]

 \draw[color=black,fill=blue!20!gray!30] (3.7,-0.7) rectangle (5.3,3.1);

\draw(0.0,0.0) node[text centered, text width=2.2cm, inner sep=0pt] (cruz1-img2) {
\includegraphics[width=1.3cm,height=1.3cm]{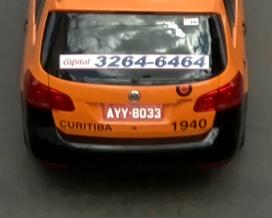}
};

\draw(1.6,0.0) node[text centered, text width=2.2cm, inner sep=0pt] (cruz1-img6) {
\includegraphics[width=1.3cm,height=1.3cm]{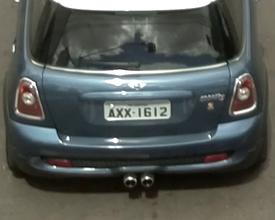}
};

\draw(4.5,0.0) node[text centered, text width=2.2cm, inner sep=0pt] (cruz1-img1) {
\frame{\includegraphics[width=1.3cm,height=1.3cm]{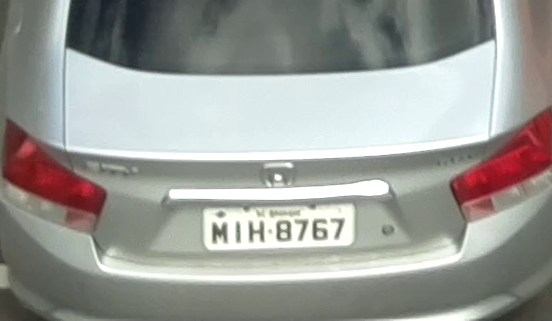}}
};

\draw(0.0,2.4) node[text centered, text width=2.2cm, inner sep=0pt] (cruz2-img1) {
\includegraphics[width=1.3cm,height=1.3cm]{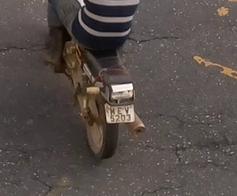}
};

\draw(1.6,2.4) node[text centered, text width=2.2cm, inner sep=0pt] (cruz2-img2) {
\includegraphics[width=1.3cm,height=1.3cm]{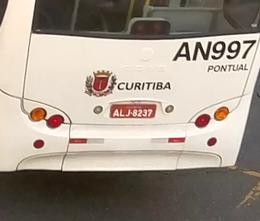}
};

\draw(4.5,2.4) node[text centered, text width=2.2cm, inner sep=0pt] (cruz2-img6) {
\frame{\includegraphics[width=1.3cm,height=1.3cm]{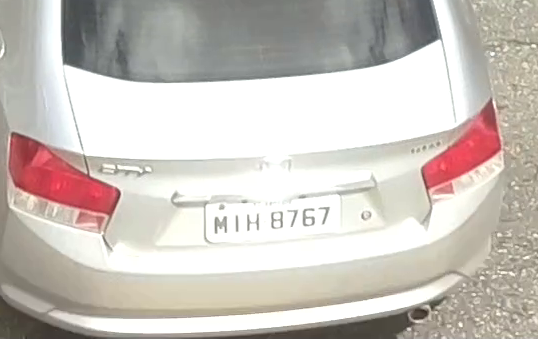}}
};

\draw (cruz1-img1.north) -- (cruz2-img1.south);
\draw (cruz1-img1.north) -- (cruz2-img2.south);

\draw (cruz1-img2.north) -- (cruz2-img1.south);
\draw (cruz1-img2.north) -- (cruz2-img2.south);
\draw (cruz1-img2.north) -- (cruz2-img6.south);

\draw (cruz1-img6.north) -- (cruz2-img1.south);
\draw (cruz1-img6.north) -- (cruz2-img2.south);
\draw (cruz1-img6.north) -- (cruz2-img6.south);

\draw [decoration={brace},decorate] (-0.8,3.2) -- (5.3,3.2) node [pos=0.5,anchor=south,yshift=0.05cm] {$n_1$ images};

\draw [decoration={brace,mirror},decorate] (-0.8,-0.8) -- (5.3,-0.8) node [pos=0.5,anchor=north,yshift=-0.05cm] {$n_2$ images};

\path[->](3.1,2.4) node[] {\Large $\dots$};
\path[->](3.1,0.0) node[] {\Large $\dots$};
\path[->](-1.0,2.4) node[rotate=90] {Camera $1$};
\path[->](-1.0,0.0) node[rotate=90] {Camera $2$};

\node[draw,align=left] at (6.5,2.2) { \small
$(n_1 \times n_2)$ - $n_1$\\
distinct pairs};

\node[] at (6.6,1.33) { \small same vehicle};
\draw[->] (4.5,1.3) -- (5.6,1.3);

\end{tikzpicture}
}
\subfigure[Matching pairs]{
\begin{tikzpicture}[scale=0.8, every node/.style={scale=0.8}]

 \draw[color=black,fill=blue!20!gray!30] (-0.8,-0.7) rectangle (0.8,3.1);

\draw(0.0,0.0) node[text centered, text width=2.2cm, inner sep=0pt] (cruz1-img1) {
\frame{\includegraphics[width=1.3cm,height=1.3cm]{./figures/two-stream/2980.png}}
};

\draw(1.6,0.0) node[text centered, text width=2.2cm, inner sep=0pt] (cruz1-img2) {
\includegraphics[width=1.3cm,height=1.3cm]{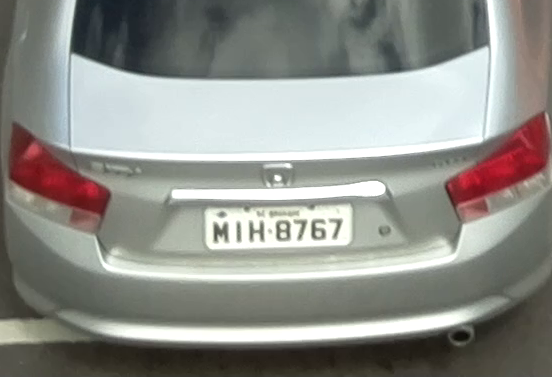}
};

\draw(4.5,0.0) node[text centered, text width=2.2cm, inner sep=0pt] (cruz1-img6) {
\includegraphics[width=1.3cm,height=1.3cm]{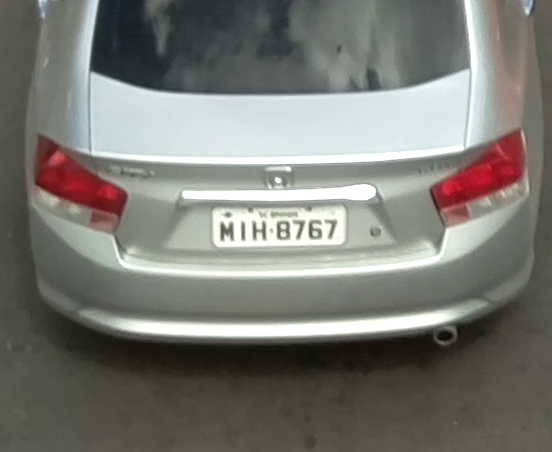}
};

\draw(0.0,2.4) node[text centered, text width=2.2cm, inner sep=0pt] (cruz2-img1) {
\frame{\includegraphics[width=1.3cm,height=1.3cm]{./figures/two-stream/741.png}}
};

\draw(1.6,2.4) node[text centered, text width=2.2cm, inner sep=0pt] (cruz2-img2) {
\includegraphics[width=1.3cm,height=1.3cm]{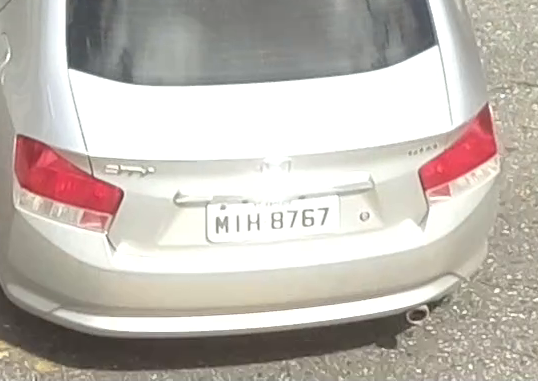}
};

\draw(4.5,2.4) node[text centered, text width=2.2cm, inner sep=0pt] (cruz2-img6) {
\includegraphics[width=1.3cm,height=1.3cm]{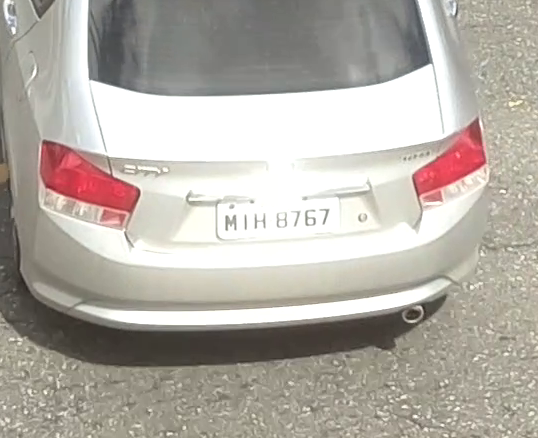}
};

\draw (cruz1-img1.north) -- (cruz2-img1.south);
\draw (cruz1-img1.north) -- (cruz2-img2.south);
\draw (cruz1-img1.north) -- (cruz2-img6.south);

\draw (cruz1-img2.north) -- (cruz2-img1.south);
\draw (cruz1-img2.north) -- (cruz2-img2.south);
\draw (cruz1-img2.north) -- (cruz2-img6.south);

\draw (cruz1-img6.north) -- (cruz2-img1.south);
\draw (cruz1-img6.north) -- (cruz2-img2.south);
\draw (cruz1-img6.north) -- (cruz2-img6.south);

\draw [decoration={brace},decorate] (-0.8,3.4) -- (5.3,3.4) node [pos=0.5,anchor=south,yshift=0.05cm] {$m$ images};

\path[->](3.1,2.4) node[] {\Large $\dots$};
\path[->](3.1,0.0) node[] {\Large $\dots$};
\path[->](-1.0,2.4) node[rotate=90] {Camera $1$};
\path[->](-1.0,0.0) node[rotate=90] {Camera $2$};

\path[->](0.0,-0.9) node[] {\small frame $j$};
\path[->](1.6,-0.9) node[] {\small frame $j$+1};
\path[->](4.5,-0.9) node[] {\small frame $j$+$m$};

\path[->](0.0,3.3) node[] {\small frame $i$};
\path[->](1.6,3.3) node[] {\small frame $i$+1};
\path[->](4.5,3.3) node[] {\small frame $i$+$m$};

\node[draw,align=left] at (6.4,1.3) { \small
All $m^2$ pairs\\
\small of each  \\
\small vehicle are\\
\small used for \\
\small training\\
\small or validation\\
\small or testing\\ 
\small (disjoint sets)};
\end{tikzpicture}
}
\vspace{-0.5mm}
\caption{\small Generation of image pairs for training, validation and testing. The same procedure is used for the license plates.}
\label{fig:pairwise}
\end{figure}

For training, evaluation and testing it is necessary to pairwise image patches. If we have $n_1$ vehicles passing through Camera~$1$ and $n_2$ vehicles passing through Camera~$2$, then we can create $n_1 \times n_2$ 
image pairs, \major{where $n_1$ is the maximum number of matching pairs and $(n_1 \times n_2) - n_1$ is the approximate number of non-matching pairs.}
Note that we have  highly imbalanced sets from non-matching pairs ($(n_1 \times n_2) - n_1)$ compared to matching pairs.
Therefore, in order to have more matching pairs, we used the MOSSE algorithm~\cite{5539960} to track a vehicle for $m$ consecutive frames, and only for the matching pairs we used all its $m$ frame occurrences to create new matching pairs. 
An advantage of using such a technique is that the object appearance in a sequence of consecutive frames usually has small image variations --~due to the vehicle motion, scene illumination changes, image noise, etc.~-- that produces distinct pairs. 
This process is depicted in Figure~\ref{fig:pairwise}. 
Using the strategy described above, we generated $5$ sets of matching/non-matching pairs, as listed in Table~\ref{tab:settings}. 

\begin{table}[!htb]
   \setlength{\tabcolsep}{12pt}
   \def\mc{\multicolumn{2}{c}}
   \def\df{\small}
   \def\de{\small}
   \def\N{\textbf}
   \renewcommand{\arraystretch}{1.0}
   \centering
   \caption{\small Number of matching/non-matching image pairs generated within each~set.}
   \vspace{1mm}
   \resizebox{0.95\linewidth}{!}{
   \begin{tabular}{ccc}
        \toprule
        Set & \# Non-matching pairs & \# Matching pairs \\  \midrule
        $01$  & $83{,}250$	& $19{,}560$  \\
        $02$  & $66{,}722$	& $17{,}370$  \\
        $03$  & $76{,}681$	& $19{,}520$  \\
        $04$  & $49{,}650$	& $14{,}177$  \\
        $05$  & $60{,}313$	& $16{,}030$  \\ \midrule
      Total & $336{,}616$   & $86{,}657$  \\ \bottomrule
   \end{tabular}
   \label{tab:settings}
   }
 \end{table}

\subsection{Experimental Setup}

\major{The CNN-OCR model was trained using the Darknet framework\footnote{\url{https://github.com/AlexeyAB/darknet/}}, while the other models were trained using Keras\footnote{\url{https://keras.io/}}. 
We performed our experiments on an Intel i7-8700K 3.7GHz CPU, 64GB RAM, with an NVIDIA Titan~Xp~GPU.}

\major{Our experiments were performed using Ubuntu~$14{.}04$,  Python $3{.}7$, OpenCV $3{.}4{.}1$, Keras $2{.}3{.}1$ and TensorFlow $1{.}15{.}2$.
All networks were trained using the Adam optimizer with a learning rate of $10$\textsuperscript{-$4$}, batch size~=~$128$, and epochs~=~$10$.
The architectures and trained models are publicly available at} \emph{\url{https://github.com/icarofua/vehicle-rear}}.


\major{We remark that we evaluated different input sizes, as well as number of filters in the convolutional layers, for both vehicle and license plate images, but better results were not~achieved.
In this sense, it is also worth noting that both models chosen by us (Small-VGG and CNN-OCR) are relatively lightweight compared to others commonly used in the literature, despite the fact that they have reached impressive results}~\cite{icaro2019,laroca2019convolutional,laroca2021towards}\major{.
More specifically, Small-VGG has $1.7$M parameters and requires $0.317$ GFLOPs, while CNN-OCR has $3.3$M parameters and requires $5.899$~GFLOPs.}

\subsection{Evaluation Metrics}

The quantitative criteria we used to assess the performance of each model are precision $P$ and  recall $R$, \minor{as defined in Equation}~(\ref{eq:metrics})
\begin{equation}
   P = \frac{tp}{tp + fp} \qquad \quad R = \frac{tp}{tp + fn}  
   \label{eq:metrics}
\end{equation}
\noindent where $tp$ denotes the number of true matchings between Cameras $1$ and $2$, $fp$ is the number of false matchings, and $fn$ the number of true matchings missed by the respective model. For  ranking  purposes,  we  also  consider  the $F$-score, which  is  the  harmonic  mean  of precision and recall, \minor{as shown in Equation}~(\ref{eq:fscore})
 \begin{equation}
   F = \frac{2}{1/P + 1/R} 
   \label{eq:fscore}
 \end{equation}  

We chose $F$-score over accuracy since the number of non-matching pairs is much larger than matching pairs and, thus, for highly imbalanced data, we can have a very low true matching rate but a very high accuracy. 

\subsection{Data Augmentation}~\label{data_aug}

For data augmentation in vehicle shape images, we used random crops between $0$ and $8$ pixels, scale between $0.8$ and $1.2$, and shear between $-8$ and $8$. 
In license plate images, we used scale between $0.8$ and $1.2$, translation between $-10$\% and $10$\%, rotation between $-5$ and $5$, and shear between $-16$ and $16$ (note that these parameter values were defined based on experiments performed in the validation set).
We used Albumentations~\cite{albumentations}, which is a well-known Python library for image augmentation, to apply these transformations.

\subsection{Ablation Study}

As shown in Table~\ref{tab:performance_shape}, we evaluated the use of several \glspl*{cnn} architectures for the identification task.
In all experiments, we used $5$ rounds of cross-validation using the $5$ sets listed in Table~\ref{tab:settings}. 
For each round, we used $2$ sets for training, $1$ for validation, and $2$ for testing.
We started with sets $01$ and $02$ for training, $03$ for validation, and $04$ and $05$ for testing; then we used $02$ and $03$ for training, $04$ for validation, and $05$ and $01$ for testing; then $03$ and $04$ for training, and so on.  Therefore, $\bar{P}, \bar{R}$ and $\bar{F}$ are the average values of precision, recall and $F$-score for these $5$~rounds. 

\begin{table}[!htb]
 \setlength{\tabcolsep}{12pt}
 \centering
 \caption{\small \minor{Vehicle identification performance based on shape:} 
 for these experiments, we evaluated several CNN architectures, exclusively based on shape features, in the Siamese Shape-Stream using different image sizes.}
 \vspace{1mm}
 \resizebox{0.99\linewidth}{!}{
 \begin{tabular}{lccc}
  \toprule
  \textbf{One-Stream (Shape-only)} & $\bar{P}$ & $\bar{R}$ & $\bar{F}$ \\
  \midrule
    Resnet$8$ ($128\times128$ px)  & $54.01$\% & $89.89$\% & $66.86$\%\\
    Lenet$5$ ($128\times128$ px)    & $89.74$\% & $71.09$\% & $78.61$\% \\
    Resnet$6$ ($128\times128$ px)  & $73.70$\% & $86.59$\% & $78.74$\%\\
    MC-CNN ($64\times64$ px)      & $83.00$\% & $82.42$\% & $82.63$\%\\
   GoogleNet ($112\times112$ px) & $79.51$\% &  $91.30$\% & $84.38$\%\\
    Matchnet ($128\times128$ px)  & $89.05$\% & $92.86$\% & $90.75$\%\\
    \textbf{Small-VGG ($\mathbf{64}\times\mathbf{64}$ px)}   & $\mathbf{90.43}$\textbf{\%} & $\mathbf{92.54}$\textbf{\%} & $\mathbf{91.35}$\textbf{\%}\\
  \bottomrule
 \end{tabular}
 }
 \label{tab:performance_shape}
\end{table}

For license plate recognition, we compared the performance of the CNN-OCR architecture against two commercial systems:
\emph{Sighthound}~\cite{masood2017sighthound} and \emph{OpenALPR}\footnote{Although OpenALPR has an open source version, the commercial version (the one used in our experiments) employs different algorithms for OCR trained with larger datasets to improve accuracy~\cite{laroca2018robust,openalprapi}.}~\cite{openalprapi}.
These systems were chosen since they are commonly used as baselines in the license plate recognition literature~\cite{goncalves2019multitask,silva2020realtime,laroca2021efficient} and also because they are robust for the detection and recognition of various license plate layouts~\cite{masood2017sighthound,openalprapi}.
It should be noted that, due to commercial reasons, little information is given about the network models used in such~systems.
As can be seen in Table~\ref{tab:ocr-results}, the CNN-OCR architecture achieved an $F$-score of $94.1$\% if we consider a perfect match (correct matching of all characters), however, if we consider partial OCR readings, then we can have an $F$-score of $97.7$\% by allowing one misreading and $98.6$\% for two misreadings. In any scenario, CNN-OCR considerably outperformed the Sighthound and OpenALPR commercial~systems.

\begin{table}[!htb]
 \setlength{\tabcolsep}{2pt}
 \def\mc{\multicolumn{3}{c}}
 \def\sc{\multicolumn{1}{c}}
 \centering
 \caption{\small \minor{Vehicle identification performance based on OCR:} comparison of the results achieved by the CNN-OCR architecture with those obtained by two well-known commercial systems. For this evaluation, we consider as true matchings the cases where exactly the same license plate characters were predicted in cameras 1 and 2. 
 }
 \vspace{1mm}
 \resizebox{0.995\columnwidth}{!}{
 \begin{tabular}{c|ccc|ccc|ccc}
  \toprule
  \sc{} & \mc{Partial matching} & \mc{Partial matching} & \mc{Perfect matching} \\
  \sc{} & \mc{2 errors} & \mc{1 error} & \mc{}  \\ \midrule
 OCR  & $\bar{P}$ & $\bar{R}$ & $\bar{F}$ & $\bar{P}$ & $\bar{R}$ & $\bar{F}$ & $\bar{P}$ & $\bar{R}$ & $\bar{F}$ \\ \midrule
  Sighthound &  $99.9${\footnotesize \%} & $84.5${\footnotesize \%} & $91.5${\footnotesize \%} & $100${\footnotesize \%} & $81.5${\footnotesize \%} & $90.0${\footnotesize \%} &  $100${\footnotesize \%} & $66.0${\footnotesize \%} & $79.3${\footnotesize \%} \\
  OpenALPR &  $99.9${\footnotesize \%} & $83.2${\footnotesize \%} & $90.7${\footnotesize \%} & $100${\footnotesize \%} & $80.4${\footnotesize \%} & $89.1${\footnotesize \%} &  $100${\footnotesize \%} & $70.0${\footnotesize \%} & $82.2${\footnotesize \%} \\
  CNN-OCR$^\ast$ &  $99.8${\footnotesize \%} & $92.0${\footnotesize \%} & $95.7${\footnotesize \%} & $100${\footnotesize \%} & $86.7${\footnotesize \%} & $92.8${\footnotesize \%} &  $100${\footnotesize \%} & $74.1${\footnotesize \%} & $84.9${\footnotesize \%} \\
  \textbf{CNN-OCR} &  $\mathbf{99.9}${\footnotesize \textbf{\%}} & $\mathbf{97.3}${\footnotesize \textbf{\%}} & $\mathbf{98.6}${\footnotesize \textbf{\%}} & $\mathbf{100}${\footnotesize \textbf{\%}} & $\mathbf{95.5}${\footnotesize \textbf{\%}} & $\mathbf{97.7}${\footnotesize \textbf{\%}} &  $\mathbf{100}${\footnotesize \textbf{\%}} & $\mathbf{88.8}${\footnotesize \textbf{\%}} & $\mathbf{94.1}${\footnotesize \textbf{\%}} \\
  \bottomrule \\[-8pt]
  \multicolumn{10}{l}{$^\ast$ CNN-OCR trained without using any images belonging to our~scenario.}
 \end{tabular}
 }
 \label{tab:ocr-results}
\end{table}

\begin{figure}[!htb]
	\centering
	\resizebox{0.9\columnwidth}{!}{
	\subfigure[\footnotesize \texttt{BAG0160}]{\includegraphics[height=6.5ex]{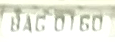}}
	\subfigure[\footnotesize \texttt{APY7367}]{\includegraphics[height=6.5ex]{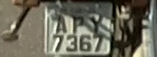}}
	\subfigure[\footnotesize \texttt{AXM6295}]{\includegraphics[height=6.5ex]{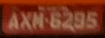}}
	} \vspace{-0.5mm}
	\resizebox{0.9\columnwidth}{!}{
	\subfigure[\footnotesize \texttt{3J66282}]{\includegraphics[height=6.5ex]{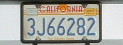}}
	\subfigure[\footnotesize \texttt{997JDG}]{\includegraphics[height=6.5ex]{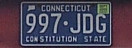}}
	\subfigure[\footnotesize \texttt{KJW5804}]{\includegraphics[height=6.5ex]{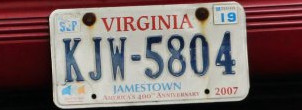}}
	} \vspace{-0.5mm} 
	\resizebox{0.9\columnwidth}{!}{
	\subfigure[\footnotesize \texttt{RK069AV}]{\includegraphics[height=6.5ex]{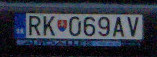}}
	\subfigure[\footnotesize \texttt{ZG4100AC}]{\includegraphics[height=6.5ex]{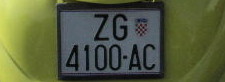}}
	\subfigure[\footnotesize \texttt{VW4X4WP}]{\includegraphics[height=6.5ex]{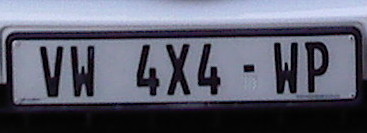}}
	}
	
	\vspace{-1mm}

	\caption{\small Examples of license plates that were correctly recognized by the CNN-OCR~architecture. The images in the first row belong to our dataset while the others belong to public datasets acquired in other~countries.}
    \label{fig:ocr-results}  
\end{figure}

It is important to highlight that we employed datasets proposed by several research groups from different countries (the same ones used by Laroca \etal~\cite{laroca2021efficient}), with only $445$ more images belonging to our scenario, to train the CNN-OCR architecture so that it is robust for various license plate layouts. In this way, as shown in Figure~\ref{fig:ocr-results}, CNN-OCR is able to correctly recognize license plates from various countries. 

As the commercial systems were not tuned specifically for our dataset/scenario, we also report in Table~\ref{tab:ocr-results} the results achieved by CNN-OCR when it was trained without using any images belonging to our~scenario. It is remarkable that CNN-OCR still outperformed both commercial systems despite the fact that they are trained in much larger private datasets, which is a great advantage, especially in deep learning-based approaches~\cite{silva2020realtime,laroca2021efficient}.
This experiment also highlights the importance of fine-tuning the CNN-OCR model to our scenario in order to achieve outstanding~results.

\begin{figure}[!htb]
	\centering
	
	\resizebox{0.9\columnwidth}{!}{
    \subfigure[\scriptsize AUF54\textcolor{red}{6}2 - AUF54\textcolor{blue}{5}2]{\includegraphics[height=6.5ex]{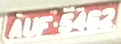}}
	\subfigure[\scriptsize AYK694\textcolor{red}{6} - AYK694\textcolor{blue}{5}]{\includegraphics[height=6.5ex]{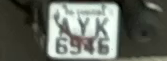}}
	\subfigure[\scriptsize A\textcolor{red}{U}H1338 - A\textcolor{blue}{D}H1338]{\includegraphics[height=6.5ex]{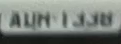}}
	}
	
	
	\resizebox{0.9\columnwidth}{!}{
	\subfigure[\scriptsize{AX\textcolor{red}{N0}937 - AX\textcolor{blue}{X8}937}]{\includegraphics[height=6.5ex]{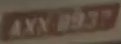}}
	\subfigure[\scriptsize{AP\textcolor{red}{I}542\textcolor{red}{4} - AP\textcolor{blue}{L}542\textcolor{blue}{3}}]{\includegraphics[height=6.5ex]{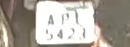}} 
	\subfigure[\scriptsize{A\textcolor{red}{OO}3681 - A\textcolor{blue}{UQ}3681}]{\includegraphics[height=6.5ex]{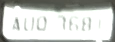}}
	} 
	
	
	\resizebox{0.9\columnwidth}{!}{
	\subfigure[\scriptsize ADS026 - ADS026\textcolor{blue}{8}]{\includegraphics[height=6.5ex]{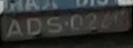}}
	\subfigure[\scriptsize A\textcolor{red}{N}A\textcolor{red}{7} - A\textcolor{blue}{O}A\textcolor{blue}{1028}]{\includegraphics[height=6.5ex]{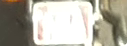}}
	\subfigure[\scriptsize \textcolor{red}{M}S938 - \textcolor{blue}{AK}S938\textcolor{blue}{3}]{\includegraphics[height=6.5ex]{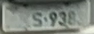}} 
	}
	
	\vspace{-0.75mm}
	
	\caption{\small Examples of license plates that were partially or not recognized by the CNN-OCR~architecture. For each license plate, we show the predicted and ground truth strings, where the red and blue characters denote the CNN-OCR misreadings and the ground truth,~respectively.}
    \label{fig:ocr-errors}  
\end{figure}

Figure~\ref{fig:ocr-errors} shows some examples in which CNN-OCR failed to correctly recognize all license plate characters. 
As can be seen, errors occur mostly due to partial occlusions, extreme light conditions, and degraded license plates.
Note that such conditions may cause one character to look very similar to another, and thus even humans can misread these license~plates (we even had to explore multiple frames and vehicle make/model information to check if the labeled string was correct in such challenging~cases).

Finally, as can be seen in Table~\ref{tab:performance_shape_ocr}, the fusion of appearance information (vehicle shape features obtained by the best network found in our experiments shown in Table~\ref{tab:performance_shape}) with textual information (\gls*{ocr}) using the proposed two-stream neural network, as described in Section~\ref{sec:method}, increased the $F$-score by nearly 5\% over each feature separately.
\begin{table}[!htb]
 \setlength{\tabcolsep}{6pt}
 \centering
 \caption{\small \minor{Vehicle identification performance based on shape and textual features}: performance of the proposed two-stream network by using the best \gls*{cnn} for shape~(Small-VGG) and the best \gls*{ocr} model~(CNN-OCR). For comparison, we included the performance of each stream when used alone.}
 \vspace{1mm}
 \resizebox{0.99\linewidth}{!}{
 \begin{tabular}{lccc}
  \toprule 
  \textbf{Architecture} & $\bar{P}$ & $\bar{R}$ & $\bar{F}$\\ \midrule
  One-Stream (Shape)  & $\phantom{0}90.43$\% & $92.54$\% & $91.35$\%\\
  One-Stream (CNN-OCR)  & $100.00$\% & $88.80$\% & $94.10$\%\\ \midrule
  \textbf{Two-Stream (Shape + CNN-OCR)}  & $\mathbf{\phantom{0}99.35}$\% & $\mathbf{98.50}$\% & $\mathbf{98.92}$\%\\
  \bottomrule
 \end{tabular}
 }
 \label{tab:performance_shape_ocr}
\end{table}

\begin{figure}[!htb]
\centering
\begin{tikzpicture}[scale=0.8, every node/.style={scale=0.8}]
\node[draw,align=left,text width=7.35cm,text height=0.2cm, inner sep=2pt] at (1.95,-2.2) {
\footnotesize Shape: \textbf{non-matching} { \normalsize \xmark} \\
\footnotesize OCR: \textbf{non-matching} { \normalsize \xmark} \\ \vspace{-3pt}
\footnotesize Two-Stream: \textbf{non-matching}  { \normalsize \xmark}};

\draw(0.0,0.0) node[] (img1) {
\frame{\includegraphics[width=3.6cm,height=3.2cm]{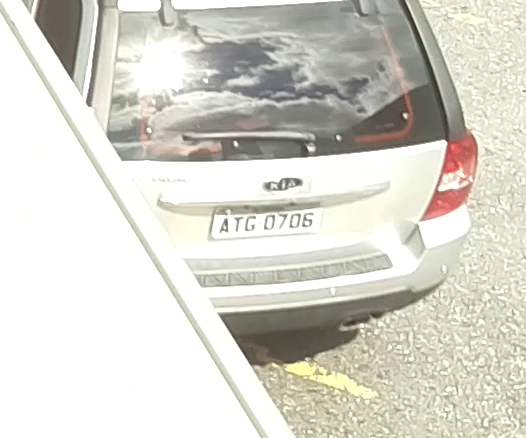}}
};

\draw(0.7,-1.0) node[] (img2) {
\frame{\includegraphics[width=2.0cm,height=1.0cm]{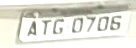}}
};

\draw(3.9,0.0) node[] (img3) {
\frame{\includegraphics[width=3.6cm,height=3.2cm]{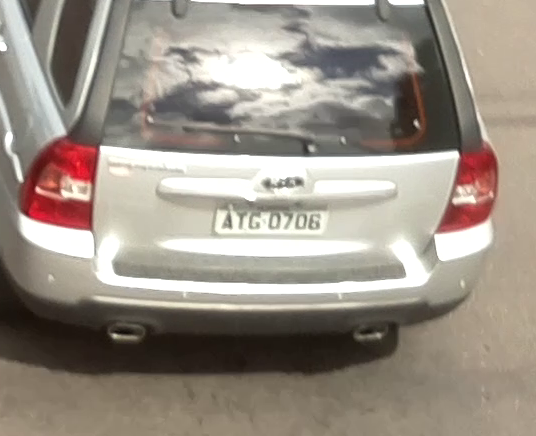}}
};

\draw(4.6,-1.0) node[] (img4) {
\frame{\includegraphics[width=2.0cm,height=1.0cm]{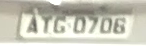}}
};
\end{tikzpicture}\vspace{0pt}
\begin{tikzpicture}[scale=0.8, every node/.style={scale=0.8}]
\node[draw,align=left,text width=7.35cm,text height=0.2cm, inner sep=2pt] at (1.95,-2.2) {
\footnotesize Shape: \textbf{matching} { \normalsize \xmark} \\
\footnotesize OCR: \textbf{non-matching} { \normalsize \cmark} \\ \vspace{-3pt}
\footnotesize Two-Stream: \textbf{non-matching}  { \normalsize \cmark}};

\draw(0.0,0.0) node[] (img1) {
\frame{\includegraphics[width=3.6cm,height=3.2cm]{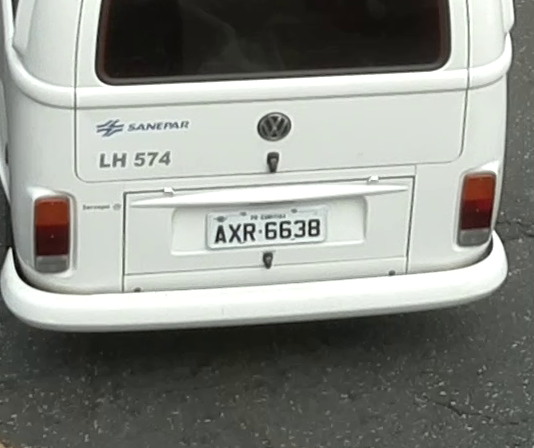}}
};

\draw(0.7,-1.0) node[] (img2) {
\frame{\includegraphics[width=2.0cm,height=1.0cm]{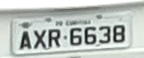}}
};

\draw(3.9,0.0) node[] (img3) {
\frame{\includegraphics[width=3.6cm,height=3.2cm]{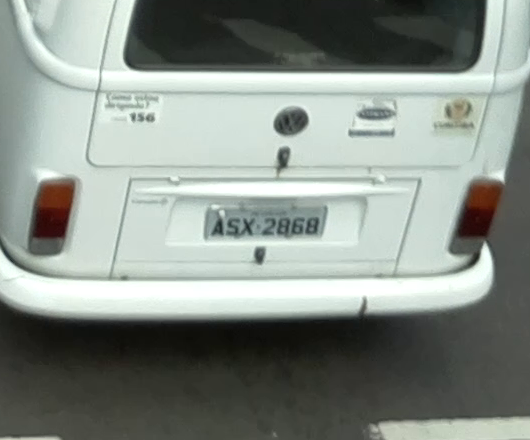}}
};

\draw(4.6,-1.0) node[] (img4) {
\frame{\includegraphics[width=2.0cm,height=1.0cm]{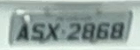}}
};
\end{tikzpicture}\vspace{0pt}
\begin{tikzpicture}[scale=0.8, every node/.style={scale=0.8}]
\node[draw,align=left,text width=7.35cm, inner sep=2pt] at (1.95,-2.2) {
\footnotesize Shape: \textbf{matching} { \normalsize \cmark} \\
\footnotesize OCR: \textbf{non-matching} { \normalsize \xmark} \\ \vspace{-3pt}
\footnotesize Two-Stream: \textbf{matching}  { \normalsize \cmark}};

\draw(0.0,0.0) node[] (img1) {
\frame{\includegraphics[width=3.6cm,height=3.2cm]{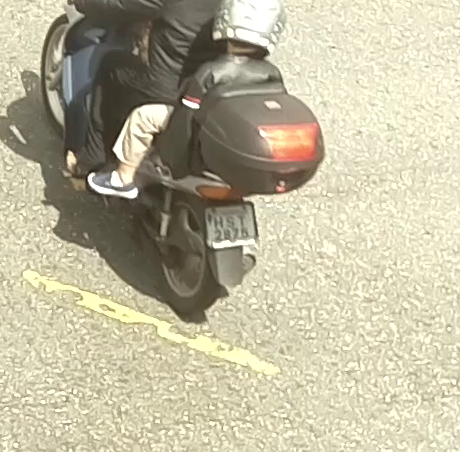}}
};

\draw(1.0,-0.85) node[] (img2) {
\frame{\includegraphics[width=1.4cm,height=1.4cm]{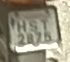}}
};

\draw(3.9,0.0) node[] (img3) {
\frame{\includegraphics[width=3.6cm,height=3.2cm]{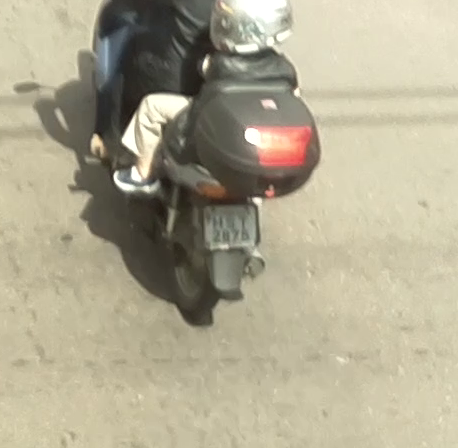}}
};

\draw(4.9,-0.85) node[] (img4) {
\frame{\includegraphics[width=1.4cm,height=1.4cm]{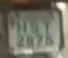}}
};
\end{tikzpicture}\vspace{0pt}
\begin{tikzpicture}[scale=0.8, every node/.style={scale=0.8}]
\node[draw,align=left,text width=7.35cm,text height=0.2cm, inner sep=2pt] at (1.95,-2.2) {
\footnotesize Shape: \textbf{non-matching} { \normalsize \cmark} \\
\footnotesize OCR: \textbf{non-matching} { \normalsize \cmark} \\ \vspace{-3pt}
\footnotesize Two-Stream: \textbf{non-matching}  { \normalsize \cmark}};

\draw(0.0,0.0) node[] (img1) {
\includegraphics[width=3.6cm,height=3.2cm]{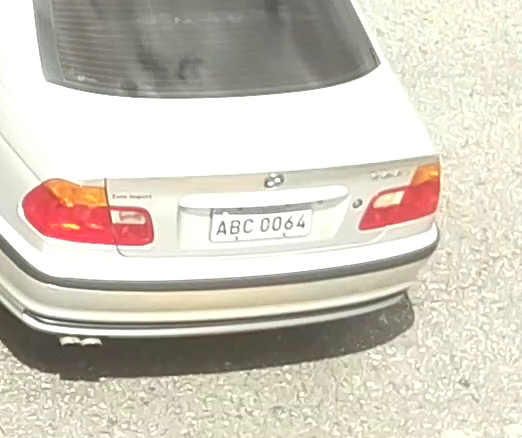}
};

\draw(0.7,-1.0) node[] (img2) {
\frame{\includegraphics[width=2.0cm,height=1.0cm]{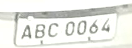}}
};

\draw(3.9,0.0) node[] (img3) {
\includegraphics[width=3.6cm,height=3.2cm]{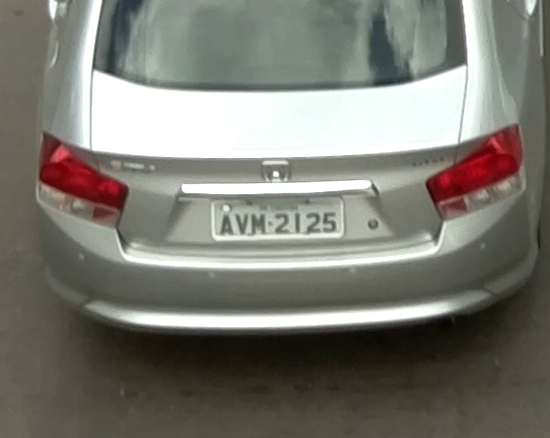}
};

\draw(4.6,-1.0) node[] (img4) {
\frame{\includegraphics[width=2.0cm,height=1.0cm]{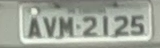}}
};
\end{tikzpicture}
\begin{tikzpicture}[scale=0.8, every node/.style={scale=0.8}]
\node[draw,align=left,text width=7.35cm,text height=0.2cm, inner sep=2pt] at (1.95,-2.2) {
\footnotesize Shape: \textbf{matching} { \normalsize \cmark} \\
\footnotesize OCR: \textbf{matching} { \normalsize \cmark} \\ \vspace{-3pt}
\footnotesize Two-Stream: \textbf{matching}  { \normalsize \cmark}};

\draw(0.0,0.0) node[] (img1) {
\includegraphics[width=3.6cm,height=3.2cm]{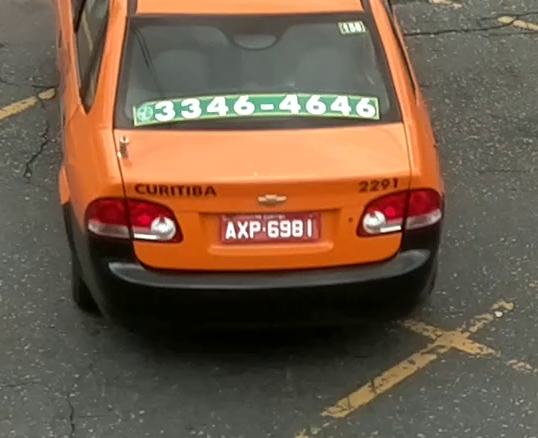}
};

\draw(0.7,-1.0) node[] (img2) {
\frame{\includegraphics[width=2.0cm,height=1.0cm]{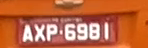}}
};

\draw(3.9,0.0) node[] (img3) {
\includegraphics[width=3.6cm,height=3.2cm]{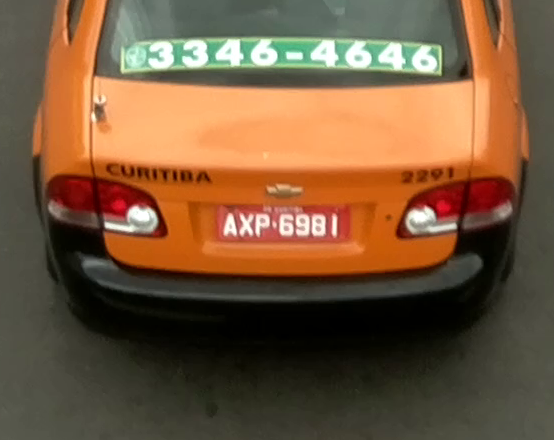}
};

\draw(4.6,-1.0) node[] (img4) {
\frame{\includegraphics[width=2.0cm,height=1.0cm]{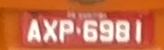}}
};
\end{tikzpicture} 
\caption{\small Inference results: the first three rows show examples where the three architectures failed: partial occlusion; CNN-Shape failed (similar shape); CNN-OCR failed (HBI-20 for the left plate, HLG-297 for the right one, while the ground truth is HST-2875). In the last two examples, all architectures found a true non-matching and a matching, respectively.}
\label{fig:results}
\end{figure}

We believe that both features have a significant level of complementarity, that is, even if CNN-OCR does not recognize all license plate characters correctly, it is still possible to correctly match the image pairs in most of the cases by using the textual and confidence information available, as well as the characters and shape similarity features.
Figure~\ref{fig:results} shows some classification results obtained by our two-stream neural~network.

As an additional contribution, we shared in GitHub\footnote{\emph{\url{https://github.com/icarofua/vehicle-rear}}} three alternative architectures that explore the same features but use additional streams and temporal~information.

\begin{table*}[!htb] 
 \setlength{\tabcolsep}{6pt}
 \centering
 \caption{\small \major{A comparison of publicly available datasets for vehicle identification with the proposed dataset called Vehicle-Rear. The entries marked with $^{\ast}$ refer to cases where only cropped patches (i.e., vehicle bounding boxes and not the entire scene) are provided by the authors.}}
 \vspace{1mm}
 \resizebox{0.99\linewidth}{!}{
 \begin{tabular}{lcccccccc}
  \toprule 
  Dataset &
  \begin{tabular}[c]{@{}c@{}}Image Resolution\end{tabular} & 
  \begin{tabular}[c]{@{}c@{}}\# Vehicles\end{tabular} &
  \begin{tabular}[c]{@{}c@{}}\# Cameras\end{tabular} &
  \begin{tabular}[c]{@{}c@{}}\# Boxes\\Labeled\end{tabular} &
  \begin{tabular}[c]{@{}c@{}} Multi-view \end{tabular} &
  \begin{tabular}[c]{@{}c@{}} Video-based \end{tabular} &
  \begin{tabular}[c]{@{}c@{}} Motorcycles \end{tabular} &
  \begin{tabular}[c]{@{}c@{}}License Plate\\Information\end{tabular} \\ \midrule
  \verisete~\cite{liu2016largescale} & \xmark$^{\ast}$ & $776$ & $20$ & $49{,}357$ & \cmark & \xmark & \xmark & \xmark \\
  \veriwild~\cite{lou2019large} & \xmark$^{\ast}$ & $40{,}671$ & $174$ & $416{,}314$ & \cmark & \xmark & \xmark & \xmark \\
  \vehicleid~\cite{liu2016deep_relative} & \xmark$^{\ast}$ & $10{,}319$ & $2$ & $90{,}000$ & \xmark & \xmark & \xmark & \xmark \\
  \citychallenge~\cite{naphade2021ai_city_challenge} & $[1280\times720]$ to $[1920\times1080]$ & $880$ & $46$ & $333{,}931$ & \cmark & \cmark & \xmark & \xmark \\
  \dataset (ours) & $[1920\times1080]$ & $2{,}093$ & $2$ & $26{,}161$ & \xmark & \cmark & \cmark & \cmark \\
  \bottomrule
 \end{tabular}
 }
 \label{tab:datasets}
\end{table*}

\section{Discussion}~\label{sec:discussion}

\major{In this section, we compare the proposed Vehicle-Rear dataset with four other well-known datasets described in the literature, namely,} \verisete~\cite{liu2016largescale}, \veriwild~\cite{lou2019large}, \vehicleid~\cite{liu2016deep_relative}, and \major{CityFlowV2}~\cite{naphade2021ai_city_challenge}\major{. An overview of these datasets is presented in Table~}\ref{tab:datasets}\major{. As can be seen, our dataset is the only one with visible/legible/labeled license plate identifiers and with all videos recorded in Full-HD resolution.
Furthermore, Vehicle-Rear and CityFlowV2 are the only datasets that provide uncropped frames, enabling the design of vehicle identification approaches that explore the entire scene.
Another point worth noting is that none of the public datasets for vehicle identification --~except ours~-- have motorcycle images, despite the fact that motorcycles are one of the most popular transportation means in
metropolitan areas, especially in developing countries~}\cite{hsu2016comparison,laroca2021efficient}\major{.
On the other hand, the images in the Vehicle-Rear dataset were not collected by as many cameras as those from CityFlowV2 and VeRi ($776$ and Wild), nor in multiple views.}

In summary, the main advantage of the proposed dataset, compared to existing ones, is that it enables the development of novel approaches/architectures for vehicle identification (both cars and motorcycles) based on the license plate identifiers in conjunction with vehicle shape~features.

\begin{figure}[!htb]
    \centering
    
    \resizebox{0.806\linewidth}{!}{ %
       \,\includegraphics[height=20ex]{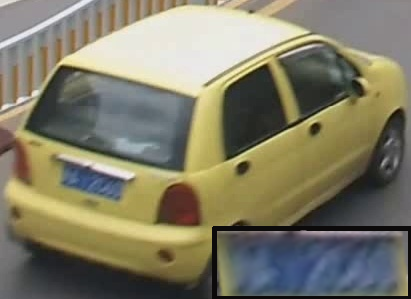}
        \includegraphics[height=20ex]{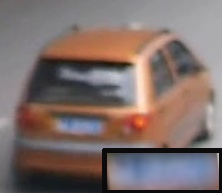}
    }
        
    \vspace{-1.1mm}    
    
    \resizebox{0.81\linewidth}{!}{ %
        \subfigure[][\verisete~\cite{liu2016largescale}\label{fig:other_datasets-veri}]{
        \includegraphics[height=23ex]{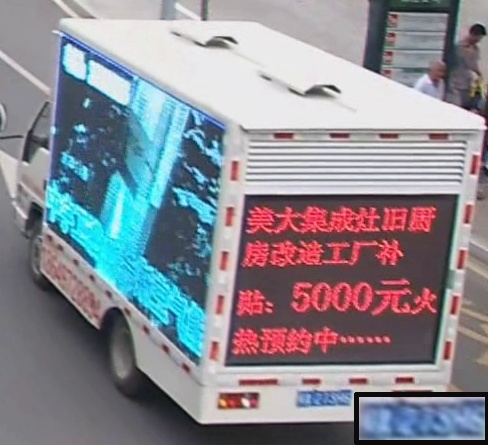}
        \includegraphics[height=23ex]{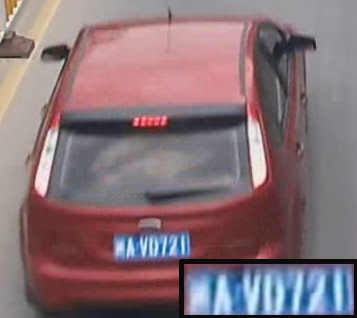}}
    }
    
    \resizebox{0.81\linewidth}{!}{ %
        \subfigure[][\veriwild~\cite{lou2019large}\label{fig:other_datasets-veri_wild}]{
        \includegraphics[height=24ex]{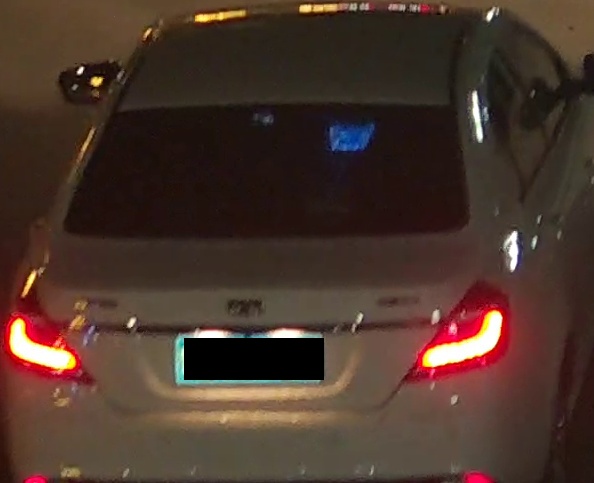}
        \includegraphics[height=24ex]{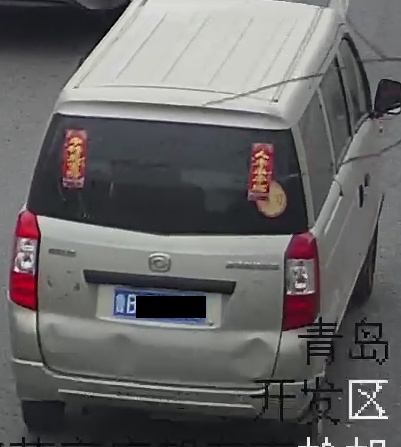}}
    }
    
    \resizebox{0.81\linewidth}{!}{ %
        \subfigure[][\vehicleid~\cite{liu2016deep_relative}\label{fig:other_datasets-vehicleID}]{
        \includegraphics[height=28ex]{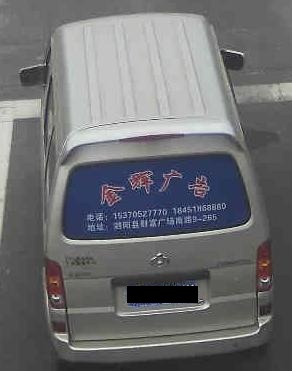}
        \includegraphics[height=28ex]{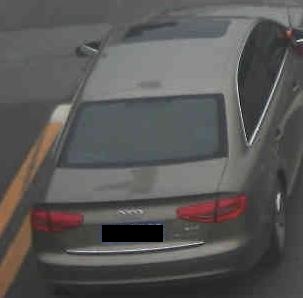}}
    }
    
    \resizebox{0.81\linewidth}{!}{ %
        \subfigure[][\citychallenge~\cite{naphade2021ai_city_challenge}\label{fig:other_datasets-cityflowv2}]{
        \includegraphics[height=16ex]{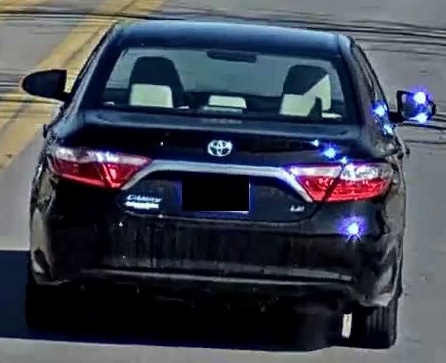}
        \includegraphics[height=16ex]{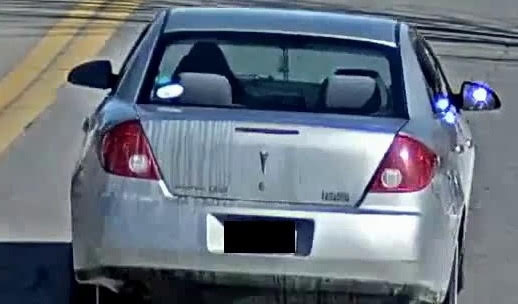}}
    }
    \vspace{-0.75mm}
    \caption{\small Vehicle rear images of \major{four} public datasets: in the \verisete dataset~(a), most license plates are not legible and the authors did not provide any annotations for the plates; in the \veriwild, \vehicleid and \major{CityFlowV2} datasets~(b-d), the license plates were redacted due to privacy~restrictions.}
    \label{fig:other_datasets}
\end{figure}

As can be seen in Figure~\ref{fig:other_datasets}, even if we consider only images from the vehicle's rear, in most of the cases the license plate identifier is illegible for the \verisete dataset, and the authors did not provide the bounding boxes and strings of the license plates in cases where they are legible, and it would be impractical (i.e., a very laborious task) to scan/label them to train/evaluate our networks.
Moreover, two state-of-the-art commercial systems that are widely employed to locate and recognize license plates from various regions/countries, Sighthound~\cite{masood2017sighthound} and OpenALPR~\cite{openalprapi}, rejected or failed in $79$\% and $96$\% of all images available in the \verisete dataset, respectively.
We emphasize that even though in~\cite{liu2016deep_learning, 8036238} the authors claimed that they extended the \verisete~dataset with license plate annotations, these annotations were not made available due to privacy restrictions~(according to the first author of~\cite{liu2016largescale, liu2016deep_learning, 8036238}). 
In the  \veriwild~\cite{lou2019large}, \major{CityFlowV2}~\cite{naphade2021ai_city_challenge} and \vehicleid~\cite{liu2016deep_relative} datasets, on the other hand, it is not even possible to exploit information from the license plate regions for vehicle identification, as they were purposely redacted in all images (with a black bounding box) by the respective authors because of privacy~restrictions.
\major{For the record, CityFlowV2 is an updated version --~with refined annotations~-- of CityFlow}~\cite{tang2019cityflow}.

In this sense, we remark that the above datasets --~as well as others available in the literature~-- have a different purpose from the one introduced in this work, as they have images from urban surveillance cameras in different resolutions and viewpoints.
As stated in~\cite{deng2021trends}, these datasets have high inter-similarity (similar visual appearance for two different makes, model and type of vehicles) and high intra-variability.

Lastly, it is important to highlight that the \dataset dataset is part of a cooperation agreement\footnote{A copy of the cooperation agreement can be obtained upon request.} between the universities involved in this project and the city where the videos were recorded. 
This agreement involves \textbf{free and open access} to the data mentioned in this~article.

\section{Conclusions}
\label{sec:conclusions}

In this paper, we introduced a novel dataset for vehicle identification that, to the best of our knowledge, is the first to consider the same camera view of most city systems used to enforce traffic laws; thus, it enables to extract features with quality and also to retrieve accurate information about each vehicle, reducing ambiguity in recognition.

To explore the \dataset dataset, we designed a two-stream CNN architecture that combines the discriminatory power of two key attributes: the vehicle appearance and license plate recognition.
For this purpose, we proposed a novel approach to compute textual similarities from a pair of license plate regions which were then combined with shape similarities extracted from a Siamese~architecture.

The proposed architecture achieved precision, recall and $F$-score values of $99.35$\%, $98.5$\%, $98.92$\%, respectively.
The combination of both features (vehicle shape and \gls*{ocr}) brought an $F$-score boost of nearly~$5$\%, solving very challenging instances of this problem such as distinct vehicles with very similar shapes or license plate identifiers.

Finally, although we achieved an $F$-score of $98.92$\% there is still room for improvement.
Some open research problems are
(i)~designing novel networks that could extract vehicle information with the same quality from even smaller image patches;
(ii)~designing a one-stream architecture that has performance comparable to multi-stream architectures; and
(iii)~exploring other fine-grained attributes or temporal sequences for vehicle~identification.
\section*{Acknowledgments}

This work was supported in part by the National Council for Scientific and Technological Development~(CNPq) under Grants 428333/2016-8,  313423/2017-2, 309292/2020-4 and 308879/2020-1, and in part by the Coordination for the Improvement of Higher Education Personnel~(CAPES) under Grant 88887.516264/2020-00.
We gratefully acknowledge the support of NVIDIA Corporation with the donation of the GPUs used for this~research.
Additionally, we thank all the support given by Curitiba's City Hall, Aditya Choudhary for his help with the code, and Diogo C. Luvizon for all his support in recording the videos used in our~experiments.
%

\scriptsize
\setlength{\bibsep}{3pt}
\bibliographystyle{IEEEtran}
\bibliography{bibtex}

\begin{thebibliography}{10}
\providecommand{\url}[1]{#1}
\csname url@samestyle\endcsname
\providecommand{\newblock}{\relax}
\providecommand{\bibinfo}[2]{#2}
\providecommand{\BIBentrySTDinterwordspacing}{\spaceskip=0pt\relax}
\providecommand{\BIBentryALTinterwordstretchfactor}{4}
\providecommand{\BIBentryALTinterwordspacing}{\spaceskip=\fontdimen2\font plus
\BIBentryALTinterwordstretchfactor\fontdimen3\font minus
  \fontdimen4\font\relax}
\providecommand{\BIBforeignlanguage}[2]{{%
\expandafter\ifx\csname l@#1\endcsname\relax
\typeout{** WARNING: IEEEtran.bst: No hyphenation pattern has been}%
\typeout{** loaded for the language `#1'. Using the pattern for}%
\typeout{** the default language instead.}%
\else
\language=\csname l@#1\endcsname
\fi
#2}}
\providecommand{\BIBdecl}{\relax}
\BIBdecl

\bibitem{bedagkargala2014survey}
A.~Bedagkar-Gala and S.~K. Shah, ``A survey of approaches and trends in person
  re-identification,'' \emph{Image and Vision Computing}, vol.~32, no.~4, pp.
  270--286, 2014.

\bibitem{8692748}
H.~{Guo}, K.~{Zhu}, M.~{Tang}, and J.~{Wang}, ``Two-level attention network
  with multi-grain ranking loss for vehicle re-identification,'' \emph{IEEE
  Transactions on Image Processing}, vol.~28, no.~9, pp. 4328--4338, Sep. 2019.

\bibitem{5763781}
W.~H. Lin and D.~Tong, ``Vehicle re-identification with dynamic time windows
  for vehicle passage time estimation,'' \emph{IEEE Trans. on Intelligent
  Transportation Systems (ITS)}, vol.~12, no.~4, pp. 1057--1063, 2011.

\bibitem{5659904}
M.~Ndoye, V.~F. Totten, J.~V. Krogmeier, and D.~M. Bullock, ``Sensing and
  signal processing for vehicle re-identification and travel time estimation,''
  \emph{IEEE Transactions on Intelligent Transportation Systems}, vol.~12,
  no.~1, pp. 119--131, 2011.

\bibitem{8265213}
Y.~Bai, Y.~Lou, F.~Gao, S.~Wang, Y.~Wu, and L.~Duan, ``Group sensitive triplet
  embedding for vehicle re-identification,'' \emph{IEEE Transactions on
  Multimedia}, pp. 2385--2399, 2018.

\bibitem{8036238}
X.~Liu, W.~Liu, T.~Mei, and H.~Ma, ``Provid: Progressive and multimodal vehicle
  reidentification for large-scale urban surveillance,'' \emph{IEEE
  Transactions on Multimedia}, vol.~20, no.~3, pp. 645--658, 2018.

\bibitem{tang2017multi}
Y.~Tang, D.~Wu, Z.~Jin, W.~Zou, and X.~Li, ``Multi-modal metric learning for
  vehicle re-identification in traffic surveillance environment,'' in
  \emph{IEEE International Conference on Image Processing}, 2017, pp.
  2254--2258.

\bibitem{lou2019large}
Y.~Lou, Y.~Bai, J.~Liu, S.~Wang, and L.-Y. Duan, ``{VERI-Wild}: A large dataset
  and a new method for vehicle re-identification in the wild,'' in \emph{IEEE
  Conf. on Computer Vision and Pattern Recognition}, 2019, pp. 3235--3243.

\bibitem{naphade2021ai_city_challenge}
M.~Naphade \emph{et~al.}, ``The 5th {AI} city challenge,'' in \emph{IEEE/CVF
  Conference on Computer Vision and Pattern Recognition Workshops}, June 2021.

\bibitem{liu2016deep_relative}
H.~Liu, Y.~Tian, Y.~Wang, L.~Pang, and T.~Huang, ``Deep relative distance
  learning: Tell the difference between similar vehicles,'' in \emph{IEEE
  Conference on Computer Vision and Pattern Recognition}, 2016, pp. 2167--2175.

\bibitem{placa_veiculo_planalto}
P.~da~República, ``{LEI N\textordmasculine{} 9.503, DE 23 DE SETEMBRO DE 1997
  - C\'{o}digo de Tr\^{a}nsito Brasileiro.}''
  \url{http://www.planalto.gov.br/ccivil_03/leis/l9503compilado.htm}, 1997,
  accessed: 2021-06-14.

\bibitem{8325486}
Y.~{Zhou}, L.~{Liu}, and L.~{Shao}, ``Vehicle re-identification by deep hidden
  multi-view inference,'' \emph{IEEE Transactions on Image Processing},
  vol.~27, no.~7, pp. 3275--3287, July 2018.

\bibitem{deng2021trends}
J.~Deng, M.~S. Khokhar, M.~U. Aftab, J.~Cai, R.~Kumar, J.~Kumar \emph{et~al.},
  ``Trends in vehicle re-identification past, present, and future: A
  comprehensive review,'' \emph{arXiv preprint arXiv:2102.09744}, 2021.

\bibitem{6875912}
B.~Tian, B.~T. Morris, M.~Tang, Y.~Liu, Y.~Yao, C.~Gou, D.~Shen, and S.~Tang,
  ``Hierarchical and networked vehicle surveillance in {ITS: A} survey,''
  \emph{IEEE Transactions on Intelligent Transportation Systems}, vol.~16,
  no.~2, pp. 557--580, 2015.

\bibitem{5983819}
R.~O. Sanchez, C.~Flores, R.~Horowitz, R.~Rajagopal, and P.~Varaiya, ``Vehicle
  re-identification using wireless magnetic sensors: Algorithm revision,
  modifications and performance analysis,'' in \emph{IEEE International
  Conference on Vehicular Electronics and Safety}, 2011, pp. 226--231.

\bibitem{6229117}
S.~Charbonnier, A.-C. Pitton, and A.~Vassilev, ``Vehicle re-identification with
  a single magnetic sensor,'' in \emph{IEEE International Instrumentation and
  Measurement Technology}, 2012, pp. 380--385.

\bibitem{Christiansen1996}
I.~Christiansen and E.~L. Hauer, ``Probing for travel time: Norway applies avi
  and wim technologies for section probe data,'' \emph{UC Berkeley
  Transportation Library}, pp. 41--44, 1996.

\bibitem{wang2019survey}
H.~Wang, J.~Hou, and N.~Chen, ``A survey of vehicle re-identification based on
  deep learning,'' \emph{IEEE Access}, vol.~7, pp. 172\,443--172\,469, 2019.

\bibitem{khan2019survey}
S.~D. Khan and H.~Ullah, ``A survey of advances in vision-based vehicle
  re-identification,'' \emph{Computer Vision and Image Understanding}, vol.
  182, pp. 50--63, 2019.

\bibitem{cabrera2011efficient}
R.~R. Cabrera, T.~Tuytelaars, and L.~Van~Gool, ``Efficient multi-camera
  detection, tracking, and identification using a shared set of
  haar-features,'' in \emph{IEEE Conference on Computer Vision and Pattern
  Recognition~(CVPR)}, 2011, pp. 65--71.

\bibitem{lowe2004sift}
D.~G. Lowe, ``Distinctive image features from scale-invariant keypoints,''
  \emph{International Journal of Computer Vision}, vol.~60, no.~2, pp. 91--110,
  2004.

\bibitem{lbp}
T.~Ojala, M.~Pietikäinen, and D.~Harwood, ``A comparative study of texture
  measures with classification based on featured distributions,'' \emph{Pattern
  Recognition}, vol.~29, no.~1, pp. 51--59, 1996.

\bibitem{MINETTO2013}
R.~Minetto, N.~Thome, M.~Cord, N.~J. Leite, and J.~Stolfi, ``{T-HOG}: {A}n
  effective gradient-based descriptor for single line text regions,''
  \emph{Pattern Recognition}, vol.~46, no.~3, pp. 1078--1090, 2013.

\bibitem{zapletal2016vehicle}
D.~Zapletal and A.~Herout, ``Vehicle re-identification for automatic video
  traffic surveillance,'' in \emph{Proceedings of the IEEE Conference on
  Computer Vision and Pattern Recognition Workshops}, 2016, pp. 25--31.

\bibitem{chen2018real}
H.~C. Chen, J.-W. Hsieh, and S.-P. Huang, ``Real-time vehicle re-identification
  system using symmelets and homs,'' in \emph{IEEE Intl. Conference on Advanced
  Video and Signal Based Surveillance}, 2018, pp. 1--6.

\bibitem{7470114}
M.~Cormier, L.~W. Sommer, and M.~Teutsch, ``Low resolution vehicle
  re-identification based on appearance features for wide area motion
  imagery,'' in \emph{IEEE Winter Applications of Computer Vision Workshops},
  2016, pp. 1--7.

\bibitem{Zhang2016}
C.~Zhang, X.~Wang, J.~Feng, Y.~Cheng, and C.~Guo, ``A car-face region-based
  image retrieval method with attention of sift features,'' \emph{Multimedia
  Tools and Applications (MTA), Springer}, pp. 1--20, 2016.

\bibitem{Luvizon:2016}
D.~C. Luvizon, B.~T. Nassu, and R.~Minetto, ``A video-based system for vehicle
  speed measurement in urban roadways,'' \emph{IEEE Transactions on Intelligent
  Transportation Systems}, vol.~18, no.~6, pp. 1393--1404, 2017.

\bibitem{10.1007/978-3-319-24261-3_7}
E.~Hoffer and N.~Ailon, ``Deep metric learning using triplet network,'' in
  \emph{Similarity-Based Pattern Recognition}.\hskip 1em plus 0.5em minus
  0.4em\relax Springer, 2015, pp. 84--92.

\bibitem{Ye:2015:ETC:2671188.2749406}
H.~Ye, Z.~Wu, R.-W. Zhao, X.~Wang, Y.-G. Jiang, and X.~Xue, ``Evaluating
  two-stream {CNN} for video classification,'' in \emph{ACM International
  Conferecen on Multimedia Retrieval}, 2015, pp. 435--442.

\bibitem{chung2017two}
D.~Chung, K.~Tahboub, and E.~J. Delp, ``A two stream siamese convolutional
  neural network for person re-identification,'' in \emph{IEEE international
  conference on computer vision}, 2017, pp. 1983--1991.

\bibitem{zagoruyko2015}
S.~Zagoruyko and N.~Komodakis, ``Learning to compare image patches via
  convolutional neural networks,'' in \emph{IEEE Conference on Computer Vision
  and Pattern Recognition}, 2015, pp. 4353--4361.

\bibitem{icaro2019}
I.~de~Oliveira, K.~V.~O. Fonseca, and R.~Minetto, ``{A} {T}wo-{S}tream
  {S}iamese neural network for vehicle re-identification by using
  non-overlapping cameras,'' in \emph{IEEE Intl. Conference on Image
  Processing}, 2019, pp. 1--4.

\bibitem{hochreiter1997lstm}
S.~Hochreiter and J.~Schmidhuber, ``Long short-term memory,'' \emph{Neural
  Computation}, vol.~9, no.~8, pp. 1735--1780, 1997.

\bibitem{8698456}
R.~{Minetto}, M.~{Pamplona Segundo}, and S.~{Sarkar}, ``Hydra: An ensemble of
  convolutional neural networks for geospatial land classification,''
  \emph{IEEE Transactions on Geoscience and Remote Sensing}, vol.~57, no.~9,
  pp. 6530--6541, 2019.

\bibitem{6165309}
S.~{Ji}, W.~{Xu}, M.~{Yang}, and K.~{Yu}, ``3d convolutional neural networks
  for human action recognition,'' \emph{IEEE Transactions on Pattern Analysis
  and Machine Intelligence}, vol.~35, no.~1, pp. 221--231, 2013.

\bibitem{Shen_2017_ICCV}
Y.~Shen, T.~Xiao, H.~Li, S.~Yi, and X.~Wang, ``Learning deep neural networks
  for vehicle re-id with visual-spatio-temporal path proposals,'' in \emph{IEEE
  International Conference on Computer Vision}, 2017, pp. 1900--1909.

\bibitem{8354181}
Y.~Zhou and L.~Shao, ``Vehicle re-identification by adversarial bi-directional
  lstm network,'' in \emph{IEEE Winter Conference on Applications of Computer
  Vision (WACV)}, vol.~00, Mar 2018, pp. 653--662.

\bibitem{liu2016deep_learning}
X.~Liu, W.~Liu, T.~Mei, and H.~Ma, ``A deep learning-based approach to
  progressive vehicle re-identification for urban surveillance,'' in
  \emph{European Conference on Computer Vision~(ECCV)}, 2016, pp. 869--884.

\bibitem{8653852}
Y.~{Lou}, Y.~{Bai}, J.~{Liu}, S.~{Wang}, and L.~{Duan}, ``Embedding adversarial
  learning for vehicle re-identification,'' \emph{IEEE Transactions on Image
  Processing}, vol.~28, no.~8, pp. 3794--3807, Aug 2019.

\bibitem{li2018reading}
H.~Li, P.~Wang, M.~You, and C.~Shen, ``Reading car license plates using deep
  neural networks,'' \emph{Image and Vision Computing}, vol.~72, pp. 14--23,
  2018.

\bibitem{dong2017cnnbased}
M.~Dong, D.~He, C.~Luo, D.~Liu, and W.~Zeng, ``A {CNN}-based approach for
  automatic license plate recognition in the wild,'' in \emph{British Machine
  Vision Conference (BMVC)}, 2017, pp. 1--12.

\bibitem{selmi2020delpdar}
Z.~Selmi, M.~B. Halima, U.~Pal, and M.~A. Alimi, ``{DELP-DAR} system for
  license plate detection and recognition,'' \emph{Pattern Recognition
  Letters}, vol. 129, pp. 213--223, 2020.

\bibitem{he2017maskrcnn}
K.~{He}, G.~{Gkioxari}, P.~{Doll{\'a}r}, and R.~{Girshick}, ``{Mask R-CNN},''
  in \emph{IEEE International Conference on Computer Vision}, Oct 2017, pp.
  2961--2969.

\bibitem{goncalves2018realtime}
G.~R. {Gon{\c{c}}alves}, M.~A. {Diniz}, R.~{Laroca}, D.~{Menotti}, and W.~R.
  {Schwartz}, ``Real-time automatic license plate recognition through deep
  multi-task networks,'' in \emph{Conference on Graphics, Patterns and Images},
  Oct 2018, pp. 110--117.

\bibitem{silva2020realtime}
S.~M. {Silva} and C.~R. {Jung}, ``Real-time license plate detection and
  recognition using deep convolutional neural networks,'' \emph{Journal of
  Visual Communication and Image Representation}, p. 102773, 2020.

\bibitem{laroca2021efficient}
R.~{Laroca}, L.~A. {Zanlorensi}, G.~R. {Gon{\c{c}}alves}, E.~{Todt}, W.~R.
  {Schwartz}, and D.~{Menotti}, ``An efficient and layout-independent automatic
  license plate recognition system based on the {YOLO} detector,'' \emph{IET
  Intelligent Transport Systems}, vol.~15, no.~4, pp. 483--503, 2021.

\bibitem{silva2017realtime}
S.~M. Silva and C.~R. Jung, ``Real-time brazilian license plate detection and
  recognition using deep convolutional neural networks,'' in \emph{Conference
  on Graphics, Patterns and Images}, Oct 2017, pp. 55--62.

\bibitem{laroca2018robust}
R.~{Laroca}, E.~{Severo}, L.~A. {Zanlorensi}, L.~S. {Oliveira}, G.~R.
  {Gon{\c{c}}alves}, W.~R. {Schwartz}, and D.~{Menotti}, ``A robust real-time
  automatic license plate recognition based on the {YOLO} detector,'' in
  \emph{International Joint Conference on Neural Networks}, 2018, pp. 1--10.

\bibitem{silva2018license}
S.~M. Silva and C.~R. Jung, ``License plate detection and recognition in
  unconstrained scenarios,'' in \emph{European Conference on Computer
  Vision~(ECCV)}, Sept 2018, pp. 593--609.

\bibitem{placa_wikipedia}
Wikipedia, ``Vehicle registration plates of {B}razil,''
  \url{https://en.m.wikipedia.org/wiki/Vehicle_registration_plates_of_Brazil},
  2021, accessed: 2021-06-14.

\bibitem{Bromley:1993:SVU:2987189.2987282}
J.~Bromley, I.~Guyon, Y.~LeCun, E.~S\"{a}ckinger, and R.~Shah, ``Signature
  verification using a ``{Siamese}'' time delay neural network,'' in
  \emph{International Conf. on Neural Information Processing Systems}, 1993,
  pp. 737--744.

\bibitem{redmon2016yolo}
J.~Redmon, S.~Divvala, R.~Girshick, and A.~Farhadi, ``You only look once:
  Unified, real-time object detection,'' in \emph{IEEE Conference on Computer
  Vision and Pattern Recognition (CVPR)}, 2016, pp. 779--788.

\bibitem{5539960}
D.~S. Bolme, J.~R. Beveridge, B.~A. Draper, and Y.~M. Lui, ``Visual object
  tracking using adaptive correlation filters,'' in \emph{IEEE Computer Vision
  and Pattern Recognition (CVPR)}, 2010, pp. 2544--2550.

\bibitem{laroca2019convolutional}
R.~{Laroca}, V.~{Barroso}, M.~A. {Diniz}, G.~R. {Gon{\c{c}}alves}, W.~R.
  {Schwartz}, and D.~{Menotti}, ``Convolutional neural networks for automatic
  meter reading,'' \emph{Journal of Electronic Imaging}, vol.~28, no.~1, p.
  013023, 2019.

\bibitem{laroca2021towards}
R.~{Laroca}, A.~B. {Araujo}, L.~A. {Zanlorensi}, E.~C. {de Almeida}, and
  D.~{Menotti}, ``Towards image-based automatic meter reading in unconstrained
  scenarios: A robust and efficient approach,'' \emph{IEEE Access}, vol.~9, pp.
  67\,569--67\,584, 2021.

\bibitem{albumentations}
A.~Buslaev, V.~I. Iglovikov, E.~Khvedchenya, A.~Parinov, M.~Druzhinin, and
  A.~A. Kalinin, ``Albumentations: Fast and flexible image augmentations,''
  \emph{Information}, vol.~11, no.~2, 2020.

\bibitem{masood2017sighthound}
S.~Z. Masood, G.~Shu, A.~Dehghan, and E.~G. Ortiz, ``License plate detection
  and recognition using deeply learned convolutional neural networks,''
  \emph{arXiv preprint arXiv:1703.07330}, 2017.

\bibitem{openalprapi}
{OpenALPR Software Solutions}, ``{OpenALPR Library \& Cloud API},''
  \url{http://www.openalpr.com/}, 2019.

\bibitem{goncalves2019multitask}
G.~R. {Gon{\c{c}}alves}, M.~A. {Diniz}, R.~{Laroca}, D.~{Menotti}, and W.~R.
  {Schwartz}, ``Multi-task learning for low-resolution license plate
  recognition,'' in \emph{Iberoamerican Congress on Pattern Recognition}, Oct
  2019, pp. 251--261.

\bibitem{liu2016largescale}
X.~{Liu}, W.~{Liu}, H.~{Ma}, and H.~{Fu}, ``Large-scale vehicle
  re-identification in urban surveillance videos,'' in \emph{IEEE International
  Conference on Multimedia and Expo}, July 2016, pp. 1--6.

\bibitem{hsu2016comparison}
G.-S.~J. Hsu and C.-W. Chiu, ``A comparison study on real-time tracking
  motorcycle license plates,'' in \emph{IEEE Image, Video, and Multidimensional
  Signal Processing Workshop}, 2016, pp. 1--5.

\bibitem{tang2019cityflow}
Z.~Tang \emph{et~al.}, ``{CityFlow}: A city-scale benchmark for multi-target
  multi-camera vehicle tracking and re-identification,'' in \emph{IEEE/CVF
  Conference on Computer Vision and Pattern Recognition}, June 2019, pp.
  8789--8798.

\end{thebibliography}

\end{document}